\pdfoutput=1

\documentclass[11pt]{article}

\usepackage[preprint]{acl}

\usepackage{times}
\usepackage{graphicx} 
\usepackage{latexsym}

\usepackage[T1]{fontenc}

\usepackage[utf8]{inputenc}

\usepackage{microtype}

\usepackage{inconsolata}

\usepackage{marvosym}
\usepackage{graphicx}

\usepackage{algorithm}
\usepackage{algcompatible}
\usepackage{booktabs}
\usepackage{multirow}
\usepackage{array}
\usepackage{diagbox}
\usepackage{amsmath}
\usepackage{xcolor}
\usepackage{subcaption}
\usepackage{amssymb}
\usepackage[T1]{fontenc}

\title{ValuePilot: A Two-Phase Framework for Value-Driven Decision-Making}

\author{
 \textbf{Yitong Luo\textsuperscript{*1,2}},
 \textbf{Hou Hei Lam\textsuperscript{2}},
 \textbf{Ziang Chen\textsuperscript{2}},
 \textbf{Zhenliang Zhang\textsuperscript{1,\Letter}},
 \textbf{Xue Feng\textsuperscript{1,\Letter}},
\\
\\
 \textsuperscript{1}State Key Laboratory of General Artificial Intelligence, BIGAI,
 \\
 \textsuperscript{2}Tsinghua University, 
 \textsuperscript{*}Work done as an intern at BIGAI
\\
\small luoyt21@mails.tsinghua.edu.cn, \{zlzhang, fengxue\}@bigai.ai
}

\begin{document}
\maketitle
\begin{abstract}

Despite recent advances in artificial intelligence (AI), it poses challenges to ensure personalized decision-making in tasks that are not considered in training datasets. To address this issue, we propose \textbf{\textit{ValuePilot}}, a two-phase value-driven decision-making framework comprising a dataset generation toolkit \textit{DGT} and a decision-making module \textit{DMM} trained on the generated data. \textit{DGT} is capable of generating scenarios based on value dimensions and closely mirroring real-world tasks, with automated filtering techniques and human curation to ensure the validity of the dataset. In the generated dataset, \textit{DMM} learns to recognize the inherent values of scenarios, computes action feasibility and navigates the trade-offs between multiple value dimensions to make personalized decisions. Extensive experiments demonstrate that, given human value preferences, our \textit{DMM} most closely aligns with human decisions, outperforming Claude-3.5-Sonnet, Gemini-2-flash, Llama-3.1-405b and GPT-4o. This research is a preliminary exploration of value-driven decision-making. We hope it will stimulate interest in value-driven decision-making and personalized decision-making within the community. 


\end{abstract}

\section{Introduction}
As artificial intelligence (AI) systems become increasingly integrated into human environments, their ability to make intelligent, adaptive, and human-like decisions has become more critical than ever \citep{lake2017building}. Traditional AI decision-making methods follow a task-oriented paradigm, optimizing for predefined objectives based on external rewards \citep{mekni2021artificial, xu2021human}. However, human decision-making follows a value-driven approach, where individuals perceive situations through their intrinsic values and choose actions that align with personal value preferences.

Value-driven decision-making is well-established in psychology and cognitive science. Schwartz’s theory of basic human values \citep{schwartz2012overview} describes a universal structure of values that influence behavior, while Maslow’s hierarchy of needs \citep{maslow1943theory} explains how human needs are prioritized in a hierarchical order, guiding decision-making based on the fulfillment of fundamental and higher-order needs. These perspectives show that human decision-making is not only about task efficiency, but also involves balancing competing values in complicated environments. By reasoning through values rather than fixed objectives, AI agents can generalize better across novel scenarios, making them more human-like in decision-making.

Despite growing interest in value alignment, existing AI approaches struggle to model individualized value-driven decision-making. Techniques such as reinforcement learning from human feedback (RLHF) \citep{christiano2017deep} and direct preference optimization (DPO) \citep{rafailov2024direct} attempt to align AI with broad human values but treat human preferences as a collective whole, failing to capture the nuanced differences in individual value preferences. Similarly, structured decision-making models like AutoPlan \citep{ouyang2023autoplan} and ReAct \citep{yao2022react} focus on optimizing for efficiency without addressing the subjective motivations that drive human decision-making. True human-like AI must not only recognize universal values but also adjust its decisions to reflect individual preferences in dynamic, real-world contexts.

Developing AI capable of value-driven decision-making requires addressing two fundamental challenges. The first challenge is recognizing value dimensions within a scenario—AI must identify which internal values are relevant to a given situation, which is an inherently subjective and context-dependent process. The second challenge is to select actions that align with value preferences. Unlike task-driven approaches, which optimize for externally defined goals, value-driven decision-making requires agents to consider both the scenario context and individual value preferences to select preferable actions. These challenges are difficult to overcome due to the lack of structured datasets that explicitly link human values and specific decisions. Existing decision-making datasets, such as ALFWorld \citep{shridhar2020alfworld} and InterCode \citep{yang2024intercode}, only focus on task completion and do not account for how human values shape action selection, making them unsuitable for training AI agents for value-driven decision-making.

To address these challenges, we propose \textbf{ValuePilot}, a two-stage framework for value-driven decision-making.
\textbf{ValuePilot} consists of two components: a structured dataset generation toolkit (\textit{DGT}) and a value-driven decision-making module (\textit{DMM}). \textit{DGT} leverages large language models (LLMs) to construct a high-quality dataset with numerical value dimension annotations, enabling agents to learn to recognize values inherent in scenarios. \textit{DMM} integrates objective action feasibility in given scenarios and the balance between multiple value preferences to make personalized decisions.

We evaluate ValuePilot across various domestic scenarios, comparing it against existing large language models and conducting a human study to assess its effectiveness. Results show that ValuePilot most closely aligns with human decisions, outperforming baseline models in selecting context-aware, preference-aligned decisions. Our framework provides a scalable solution for training AI agents capable of making generalizable, interpretable, and personalized decisions.

\section{Value-Driven Dataset Generation}
\begin{figure*}
    \centering
    \includegraphics[width=1\linewidth]{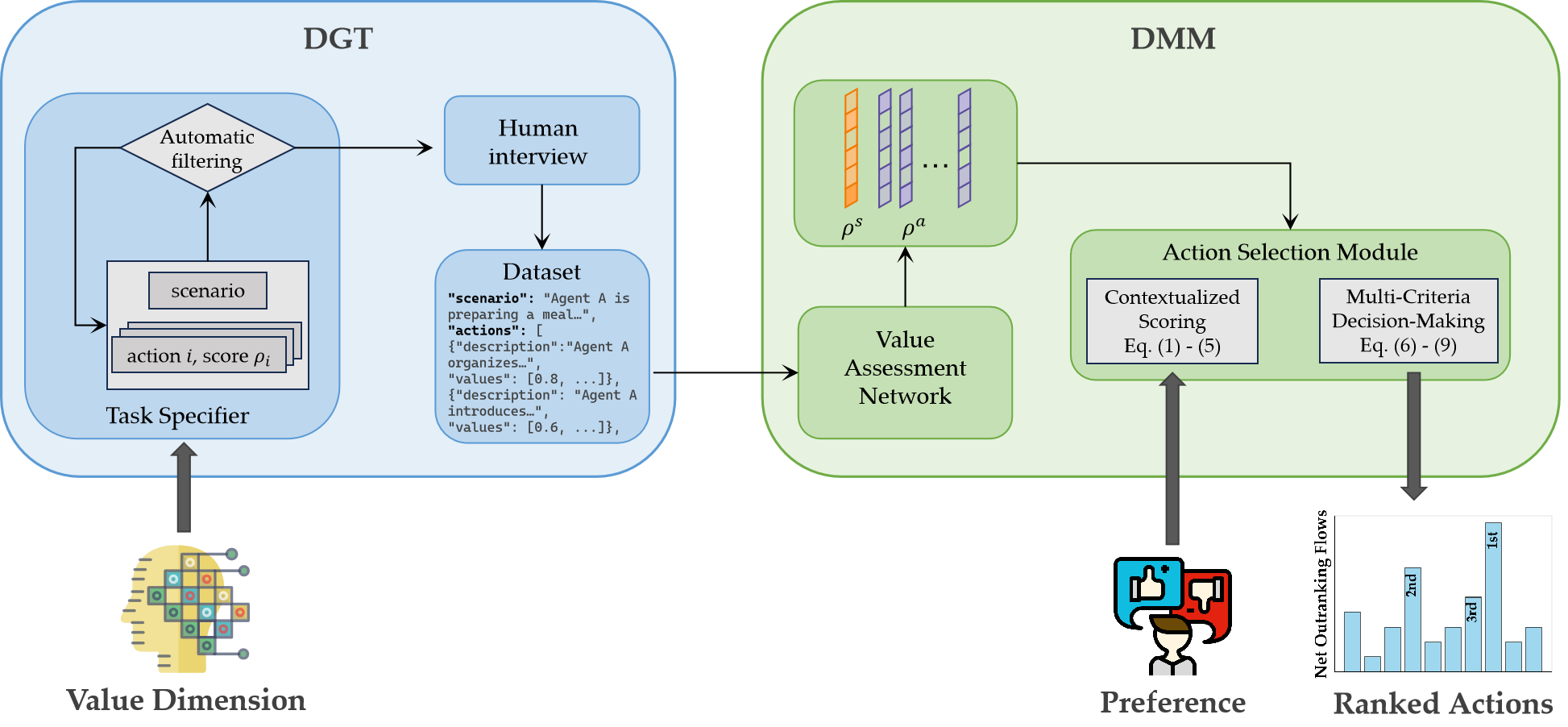}
    \caption{ValuePilot framework. The ValuePilot framework simulates individual preferences and guides AI decision-making through a two-phase process. a) \textbf{\textit{DGT}}. A dataset generation toolkit using large language model to generate structured dataset following detailed instructions which includes scenario descriptions, potential actions, and corresponding value dimension scores. b) \textbf{\textit{DMM}}. It consists of two modules: Value Assessment Network and the Action Choosing Module.The \textbf{Value Assessment Network} processes scene and action representations through an encoder, followed by a multi-head self-attention mechanism and average pooling, finally mapping the embeddings into the objective value scores of scenarios and actions via an MLP. The \textbf{Action Selection Module} receives the rating output from the Value Assessment Network, combined with pre-personalized value preferences. Through a purely mathematical white-box process involving Contextualized Scoring and the PROMETHEE method, it produces the final ranked actions.}
    \label{fig:pipeline}
    \vspace{-14pt}
\end{figure*}

To systematically construct datasets that reflect value-driven decision-making, we introduce \textit{DGT} (ValuePilot Dataset Generation Toolkit). As shown in \autoref{fig:pipeline}, \textit{DGT} generates scenarios, a set of possible actions with scores in each value dimension, ensuring a structured and interpretable dataset for training AI agents to recognize the inherent values of scenarios.

\subsection{Task Specifier}
The Task Specifier is the core component of our automated value-driven dataset generation toolkit. It processes input value dimensions and their detailed descriptions, restructures them into a structured, high-quality prompt for GPT-4, ensuring that the generated scenarios, actions, and scores align with the selected value dimensions. This structured automation enables efficient, large-scale dataset generation while ensuring consistency in format and quality. Its modular design allows for seamless adaptation to new value dimensions and decision contexts, making it highly scalable and flexible for diverse AI applications.

\paragraph{Scenario Generation}
GPT-4 generates scenarios based on selected value dimensions, ensuring they play a crucial role while avoiding explicit mention of value-related terms. For example, if `curiosity' and `safety' are chosen, the scenario might involve an agent deciding whether to explore an unknown area, balancing potential discovery (curiosity) against risk factors (safety).

\paragraph{Action Listing and Scoring}
For each scenario, GPT-4 generates a set of ten actions, each designed to reflect different trade-offs between all selected value dimensions. The actions are not only tailored to fit the scenario but also structured to span a range of responses, ensuring diversity in decision-making approaches.
Each action is then scored on a continuous scale from -1 to 1 along each value dimension, where -1 indicates a significant contradiction to the value (e.g., a risky action greatly reduces safety), 0 represents unrelated, and 1 signifies strong alignment with the value (e.g., an exploratory action highly reinforces curiosity).

\paragraph{Automatic Filtering}
After the initial scenario and action generation, we instruct the language models to `forge' the initially used value dimensions. During the subsequent review phase, the models must reassess and re-identify the value dimensions based purely on the text descriptions of the scenarios and actions, without any preconceived biases. This self-filtering approach helps to eliminate scenarios and actions that involve other, irrelevant value dimensions, ensuring that each generated scenario and corresponding action set include only mentioned values.

\subsection{Human Review}
Each dataset then undergoes human review to ensure scenario-action coherence and accurate value alignment. Scores are reviewed for consistency, and extreme or unrealistic actions are filtered out to maintain dataset reliability.

This step is critical because training models on synthetically generated data can lead to a degradation in model performance over time \cite{shumailov2024ai}. By manually curating the data, we mitigate these risks and enhance the dataset's utility for training robust models, ensuring that the data remains diverse, relevant, and of high quality.


\section{Value-Driven Decision Making}
We develop a value-based action selection framework that leverages the dataset to train a language model capable of value-driven decision-making. \textit{DMM} (ValuePilot Decision-making Module), shown in \autoref{fig:pipeline}, aligns an agent’s actions with human-like values, such as curiosity, fairness, and safety, which allows it to make decisions that reflect individual characteristics rather than generic, goal-oriented strategies.

\subsection{Value Assessment Network}
First, we employ the T5 (Text-to-Text Transfer Transformer) \citep{raffel2020exploring} base model’s encoder to convert textual descriptions of scenarios and actions into embeddings, providing a foundation for training the Value Assessment Network. Compared to larger models, T5 provides a lightweight and efficient architecture, making it suitable for real-world applications where computational efficiency is crucial. Our experiments comparing T5 with other language models demonstrate that T5 achieves competitive performance while significantly reducing computational costs, further validating its suitability for this task.

The Value Assessment Network evaluates potential actions by estimating their impact on each value dimension. It generates a vector whose length corresponds to the number of value dimensions, with each element representing the degree to which an action positively or negatively influences this value. To achieve this, the network processes textual representations of the scene and action set through an encoder, which converts them into a hidden state of shape \(H \times L\times b\) (encoder last hidden state length, sequence length, batch size). A multi-head self-attention mechanism with four attention heads then extracts meaningful relationships between the scenario and the available actions, allowing the model to attend to multiple contextual cues simultaneously. The resulting sequence embeddings are averaged across the sequence length \(L\) to produce a compact representation, which is then fed into a Multi-Layer Perceptron (MLP) with two fully connected layers (hidden size 128). This final transformation maps the dense vector into the output space. A tanh activation function is applied to the output, ensuring that scores range between -1 (negative impact) and 1 (positive impact), thereby enabling a structured evaluation of how each action influences different value dimensions.

\subsection{Action Selection Module}
In decision-making, selecting an appropriate action requires considering three key elements: the action itself, the scenario in which it takes place, and the individual's value preferences. An action cannot be evaluated in isolation—its feasibility and desirability depend on the scenario context, while an individual's value preferences further shape the selection process. However, directly integrating these factors is challenging, as preferences influence decision-making in nonlinear and context-dependent ways.

To address this, the Action Selection Module processes the output from the Value Assessment Network, combined with predefined value preferences, to compute a final ranking of available actions. It consists of two key stages: \textbf{Contextualized Scoring} and \textbf{Action Ranking Using PROMETHEE} (Preference Ranking Organization Method for Enrichment Evaluations \citep{brans1985note}). In the first stage, it computes both objective and subjective scores for each action, incorporating scenario context and individual preferences to refine decision impact. In the second stage, it ranks actions using a structured multi-criteria decision analysis approach, ensuring that the selected action aligns with personal biases. Details are shown in Algorithm \ref{alg:action-selection} in Appendix.

\subsubsection{Contextualized Scoring}
Humans tend to exhibit a bias toward values that align with their preferences, often assigning higher subjective scores to such values while avoiding extreme ratings and favoring moderate ones instead. To preprocess the value preferences \( p \), we apply a sigmoid function to emphasize extremes:
\vspace{-1mm}
\begin{align}
    p' = \frac{1}{1 + e^{-(p - 0.5) \times 10}}
\end{align}
This transformation keeps moderate values near 0.5 while pushing distinct values toward the extremes.

Each action is evaluated based on two components: an \textbf{objective score} derived directly from the Value Assessment Network and a \textbf{preference discrepancy score} that adjusts for individual tendencies. The objective score represents how well an action aligns with a given value dimension in a general sense, while the preference discrepancy score captures how much this alignment deviates from the individual’s expected preference. Formally, for each value dimension \( j \), the objective scores for a scenario and an action are \( \rho^s_j \) and \( \rho^{a_i}_j \) , which are direct outputs of the Value Assessment Network for the scenario and action, respectively.

To incorporate subjective biases, we define a scenario preference discrepancy score (\( d^s_j \)) and an action preference discrepancy score (\( d^{a_i}_{j} \)), which measure how much the model’s predicted score deviates from the individual’s preference:
\vspace{-1mm}
\begin{equation}
    \begin{aligned}
        d^s_j &= 1 - \left| \left| \rho^s_j \right| - p'_j \right| \\
        d^{a_i}_{j} &= 1 - \left| \left| \rho^{a_i}_j \right| - p'_{j} \right|
    \end{aligned}
\end{equation}

These scores are higher when the model’s predictions align closely with individual preferences and lower when there is a large discrepancy, reflecting a tendency to avoid mismatched actions or scenarios.

The final integrated scores for a scenario and action are computed by combining objective and subjective components using a weighted sum:
\vspace{-1mm}
\begin{equation}
    \begin{aligned}
    r^s_j &= (d^s_j \times w_s) + (\rho^s_j \times (1 - w_s)) \label{eq:4}\\
    r^{a_i}_j &= (d^{a_i}_j \times w_a) + (\rho^{a_i}_j \times (1 - w_a))
    \end{aligned}
\end{equation}

where \( w_s \) and \( w_a \) control the weight of subjective versus objective factors in decision-making. Empirical testing suggests that 
\( w_s =w_a=0.3\) best models human decision patterns.

Finally, the adjusted scenario score is used as a scaling coefficient to ensure that an action is evaluated within the context of its scenario, rather than in isolation:
\vspace{-1mm}
\begin{equation}
    r_{i,j} = \frac{1}{1 + e^{-\left| r_j^s \right| }} \times r_{j}^{a_i} \label{eq:5}
\end{equation}

This formulation ensures that actions are weighted not only by their intrinsic value alignment but also by their appropriateness within a given scenario, leading to more interpretable, context-sensitive decision-making.

\subsubsection{Action Ranking Using PROMETHEE}
The action ranking problem in our framework is inherently a multi-criteria decision-making (MCDM) task, as each action must be evaluated across multiple value dimensions, each of which may contribute differently to the final decision. Unlike simple ranking approaches that select the action with the highest single score, our method must balance trade-offs across conflicting value dimensions while incorporating individual preferences.

To achieve this, we apply the PROMETHEE, a widely used MCDM approach. PROMETHEE allows for pairwise comparisons between actions based on their relative preference across multiple criteria, making it well-suited for our task, where actions may excel in one value dimension while underperforming in another. The pairwise preference score for two actions 
\(i\) and \(i'\) under value dimension \(j\) is computed as:
\vspace{-2mm}
\begin{equation}
V_{ii',j} = \frac{1}{1 + e^{-(r_{i,j} - r_{i',j})}}
\end{equation}

This enhances sensitivity to small differences in scores, allowing for fine-grained preference ranking. The overall preference score for an action is then computed as:
\vspace{-2mm}
\begin{equation}
\Tilde{V_{ii'}} = \sum_{j=1}^{m} p'_j \cdot V_{ii',j} \label{eq:7}
\end{equation}

which weights each value dimension according to individual value preferences.

For each action \(i\), we compute:
\vspace{-1mm}
\begin{itemize}
    \item Positive Flow (\(\phi^+_i\)): The average dominance of action \(i\) over others.
    \vspace{-1mm}
    \item Negative Flow (\(\phi^-_i\)): The average weakness of action \(i\) compared to others.
\end{itemize}
\vspace{-3mm}
\begin{align}
\phi^+_i = \frac{1}{n} \sum_{i'\neq i}^{n} \Tilde{V_{ii'}} \quad
\phi^-_i = \frac{1}{n} \sum_{i'\neq i}^{n} \Tilde{V_{i'i}}
\end{align}
\vspace{-1mm}
The final ranking is determined by the net outranking flow:
\vspace{-2mm}
\begin{equation}
\phi_i = \phi^+_i - \phi^-_i
\end{equation}

where a higher net flow indicates a more preferable action. The action with the highest net flow is selected as the optimal choice within the scenario, as detailed in Algorithm \ref{alg:action-selection} in Appendix.

\section{Experiments}

\subsection{Dataset Preparation}
We select 6 value dimensions that are very important in the daily life of human beings - \textbf{Curiosity}, \textbf{Energy}, \textbf{Security}, \textbf{Happiness}, \textbf{Intimacy}, and \textbf{Fairness}. These values are carefully chosen based on a review of relevant literatures \citep{maslow1943theory, qiu2022valuenet}.
They are distinct and representative, capturing a broad range of individual and multi-agent value considerations.
Utilizing \textit{DGT}, we generate the dataset comprising \textbf{11,938 scenarios} and a total of \textbf{100,255 actions}.
A detailed description of each value dimension, a discussion of why these six value dimensions, and the statistics of the dataset can be found in Appendix~\ref{appendix:dataset}.

\subsection{Value recognition}
This subsection conducts experiments to test \textbf{ValuePilot}'s performance in recognizing values from scenarios, which is the first critical challenge in value-driven decision-making. Implementation details are included in Appendix \ref{appendix:dataset}.

\subsubsection{Baselines and Evaluation Metrics}
We compare our method with open-source LLMs, including llama-3.5-70b, llama-3.5-405b, mixtral-8x22b, and gemini-1.5-flash.
Two rigorously designed metrics are employed to comprehensively assess model performance from complementary perspectives. First, we introduce a threshold-based accuracy metric, inspired by the Fault Tolerant \cite{lee1990fault} concept in engineering, which quantifies classification performance across six ordinal value dimensions. This metric computes dimension-wise accuracy using absolute error thresholds of \( t = 0.2 \) (practical alignment) and \( t = 0.05 \) (strict alignment), subsequently aggregating results across all dimensions and samples through hierarchical averaging. Formally, for \( N_t \) test samples (where \( N_t \in \mathbb{N}^* \) denotes the cardinality of the test set) and \( D = 6 \) dimensions (with \( D \) indexing the value categories \( \mathcal{V} = \{v_1,...,v_6\} \)), the average accuracy is defined as:
\vspace{-2mm}
\begin{equation}
\text{AvgAcc} = \frac{1}{N_tD} \sum_{i=1}^{N_t}\sum_{d=1}^D \mathbb{I}\big(|\hat{y}_d^{(i)} - y_d^{(i)}| < t\big),
\end{equation}
where \( \mathbb{I}: \mathbb{R} \to \{0,1\} \) is the indicator function.

This dual-threshold approach addresses both practical deployment requirements and precise value alignment challenges. Second, we adopt the mean absolute error (MAE) as a continuous evaluation metric to complement threshold-based measurements. The MAE calculation follows:
\vspace{-2mm}
\begin{equation}
\text{MAE} = \frac{1}{N_tD}\sum_{i=1}^{N_t}\sum_{d=1}^D |\hat{y}_d^{(i)} - y_d^{(i)}|
\end{equation}

Although threshold accuracy provides intuitive categorical assessments, it risks being overly lenient for borderline cases (e.g., predictions of 0.18 vs. 0.22 when t=0.2), which MAE penalizes proportionally.

\subsubsection{Results}
As shown in \autoref{tab:performance}, our Value Assessment Network outperforms other open-source LLMs across all evaluation metrics. The model achieves 66.70\% average accuracy at \( t = 0.2 \) and 40.00\% at \( t = 0.05 \), representing respective improvements of 15.09 and 14.36 percentage points over the strongest baseline (Gemini-1.5-Flash). The MAE of 0.19 constitutes a 36.7\% relative error reduction compared to baseline models.

Overall, we find that: first, the performance gap widens substantially under stricter evaluation thresholds, suggesting our model better captures subtle value distinctions that baseline LLMs fail to discern; second, all baseline models exhibit comparable performance ranges (41\% to 52\% accuracy at \( t = 0.2 \)), highlighting fundamental limitations in current LLMs' zero-shot value alignment capabilities. 

\begin{table}[ht]
\centering
\caption{Comparison between our method and baselines in terms of their ability to recognize inherent values of scenarios. Our method outperforms open-source LLMs.}
\label{tab:performance}
\begin{tabular}{@{}cccc@{}}
\toprule
\textbf{Model}             & \multicolumn{2}{c}{\textbf{AvgAcc (\%)}} & \textbf{MAE} \\ 
                          & $t = 0.2$ & $t = 0.05$ &              \\ \midrule
llama-3.5-70b             & 40.90          & 17.74          & 0.30         \\ 
llama-3.5-405b            & 41.62          & 18.00          & 0.29         \\ 
mixtral-8x22b& 42.71          & 18.39          & 0.29         \\ 
gemini-1.5-flash          & 51.61          & 25.64          & 0.24         \\ 
\multirow{2}{*}{\shortstack{\textbf{Value Assessment} \\ \textbf{Network (Ours)}}}& \multirow{2}{*}{\textbf{66.70}} & \multirow{2}{*}{\textbf{40.00}} & \multirow{2}{*}{\textbf{0.19}}         \\ \\ \bottomrule
\end{tabular}
\end{table}

\subsection{Value-driven Decision-making}
To test \textbf{ValuePilot}'s performance in human-like decision-making, this subsection conducts human study to collect self-reported value preferences and corresponding decisions in domestic scenarios. 
Specifically, we designed a questionnaire with 11 formal questions to collect data from 40 subjects. Each subject rated the importance of six value dimensions in their daily lives on a scale from 0 to 1. To ensure understanding, a pilot study with three preliminary questions was conducted, allowing subjects to familiarize themselves with the process. After that, subjects rescaled the importance of the value dimensions, producing a final scoring vector. They were then presented with scenarios and action lists, ranking their willingness to choose actions based on their value tendencies. This process collects a six-dimensional value vector and corresponding action ranking data for each subject.
Then, we give the collected value preferences to \textbf{ValuePilot} and the baselines to compute the similarity between their decisions and human decisions.

The baselines include mainstream LLMs: Claude-3.5-sonnet \citep{anthropic2024claude}, Llama-3.1-70b \citep{dubey2024llama}, Llama-3.1-405b, Gemini-1.5-pro \citep{team2023gemini}, Gemini-2-flash, GPT-4o \citep{achiam2023gpt} and GPT-4o-mini. Detailed experiment setups and demographic analysis are provided in Appendix~\ref{appendix:Demographic}.


\subsubsection{Evaluation Metric}
We use two complementary metrics to comprehensively assess the alignment with human decision patterns. The primary metric, termed \textbf{Order-Sensitive Similarity (OS-Sim)}, extends the Jaccard index through progressive prefix comparison. Let \( i \in \{1, \dots, N_{q}\} \) represent the question sample in the questionnaire. For two sequences of actions \( S^{(i)} = [s_1^{(i)}, s_2^{(i)}, \ldots, s_n^{(i)}] \) and \( T^{(i)} = [t_1^{(i)}, t_2^{(i)}, \ldots, t_n^{(i)}] \), define \( S_d^{(i)} \triangleq \{s_1^{(i)}, \ldots, s_d^{(i)}\} \) and \( T_d^{(i)} \triangleq \{t_1^{(i)}, \ldots, t_d^{(i)}\} \). The similarity at depth \( d \in \{1, \ldots, n\} \) is a modified Jaccard Similarity \(\text{sim}_d^{(i)}(S^{(i)}, T^{(i)}) = \frac{|S_d^{(i)} \cap T_d^{(i)}|}{d}\).   
The mean $\text{OS-Sim}$ is
\vspace{-2.5mm}
\begin{equation}
    \text{OS-Sim} = \frac{1}{N_q} \sum_{i=1}^{N_q} \text{OS-Sim}^{(i)},
\end{equation}
where \(\text{OS-Sim}^{(i)}(S^{(i)}, T^{(i)}) = \frac{1}{n} \sum_{d=1}^n \text{sim}_d^{(i)}\).
This design captures both immediate preference alignment in early positions and gradual decision pattern matching throughout the sequence. Detailed examples and the advantages of Os-Sim over methods such as \textit{Spearman Rank Correlation} or \textit{Kendall Tau} are provided in Appendix~\ref{appendix:examples_os_sim}.



By applying this method, we can objectively compare the ranking lists generated by our model and the LLMs with the actual lists provided by the participants, allowing for a robust evaluation of model performance.

To address real-world constraints where only the primary action selection matters, we introduce \textbf{First-Action Accuracy} as a critical secondary metric. This measure quantifies the percentage of trials where the model's top-ranked action matches the human subject's first choice:
\vspace{-2mm}
\begin{equation}
\text{First-Acc} = \frac{1}{N_{q}}\sum_{i=1}^{N_{q}} \mathbb{I}(s_1^{(i)} = t_1^{(i)}),
\end{equation}
where $\mathbb{I}$ denotes the indicator function. This dual-metric approach enables nuanced evaluation of both comprehensive sequence alignment and practical decision-making accuracy.

\subsubsection{Results}
\begin{figure}[!htp]
    \centering
    \includegraphics[width=1\linewidth]{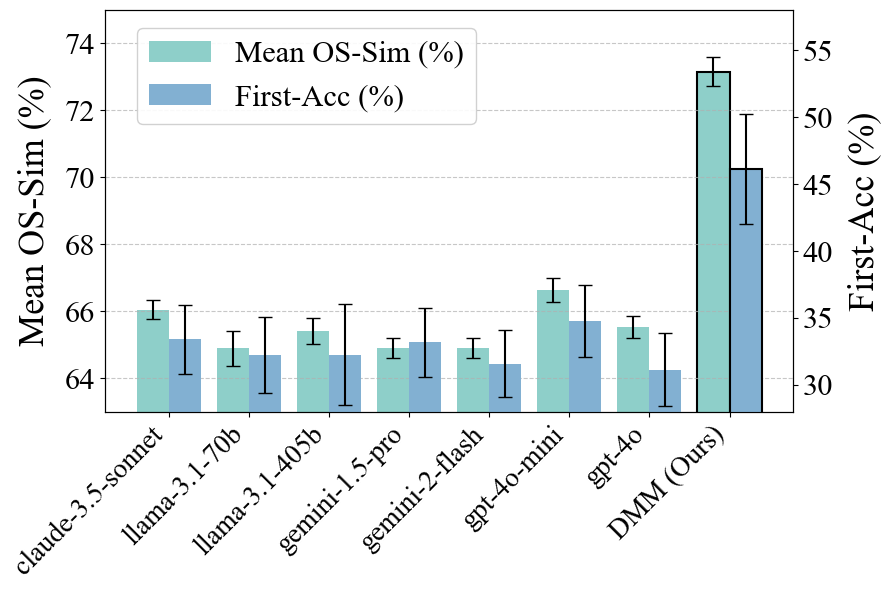}
    \caption{Comparison between our DMM and LLMs in terms of their decision rankings' similarity to human decisions. DMM achieves the highest scores, outperforming the best baseline (GPT-4o-mini) by +6.53\% in OS-Sim and +11.37\% in First-Acc.}
    \label{fig:similarity}
\end{figure}

As shown in \autoref{fig:similarity}, our DMM framework demonstrates significant superiority over state-of-the-art LLMs in both sequence-level alignment and critical first-action prediction accuracy.
The proposed OS-Sim metric reveals that our model achieves a mean alignment score of $73.16\% (\pm 0.43\%)$, outperforming the strongest LLM baseline (GPT-4o Mini at $66.63\% \pm 0.35\%$). This substantial improvement indicates enhanced capability in capturing both the hierarchical preference structure and nuanced position-sensitive choices inherent in human decision sequences.

Notably, our framework exhibits particularly strong performance in the pragmatically crucial \textbf{First-Action Accuracy} metric, achieving $46.14\% (\pm 4.09\%)$ prediction accuracy for initial human choices. This represents a $11.37\%$ relative improvement over the best-performing LLM (GPT-4o Mini at $34.77\% \pm 2.67\%$), demonstrating exceptional fidelity in modeling primary decision determinants. The performance differential exceeds conventional benchmarks in behavioral alignment tasks, suggesting that explicit value prioritization modeling provides distinct advantages over implicit pattern learning in LLMs.

\begin{table}[ht]
\centering
\caption{Ablation Study for DMM. }
\label{tab:ablation}
\begin{tabular}{@{}ccc@{}}
\toprule
\multirow{2}{*}{\textbf{Model}}& \multirow{2}{*}{\shortstack{\textbf{Mean} \\ \textbf{OS-Sim (\%)}}} & \multirow{2}{*}{\textbf{First-Acc (\%)}}
\\ \\ \midrule
Only Action                         & \(60.23 \pm 0.17\)                              & \(32.27 \pm 0.58\)                                 \\
w/o Preference                  & \(61.07 \pm 0.29\)                              & \(31.82 \pm 0.94\)                                 \\
w/o Subjective                 & \(68.93 \pm 0.49\)                              & \(43.45 \pm 1.16\)                                 \\
w/o Scenario                 & \(69.99 \pm 0.57\)                               & \(43.64 \pm 1.32\)                                 \\
\multirow{2}{*}{\shortstack{\textbf{DMM} \\ \textbf{(Full Version)}}}                        & \multirow{2}{*}{$\mathbf{73.16 \pm 0.43}$}                              & \multirow{2}{*}{\(\mathbf{46.14 \pm 4.09}\)}                           \\      \\ \bottomrule
\end{tabular}
\end{table}

\subsection{Ablation Study}
To assess the contribution of key components in our decision-making framework, we conduct an ablation study, systematically removing or modifying different elements of our model and evaluating its performance on two metrics.

We evaluate the full DMM model against four ablated variants, each removing one or all crucial components to analyze its impact.


\begin{enumerate}
    \item \textbf{Only Action}: The model ranks actions without considering scenario context or individual value preferences, relying solely on action scores from the Value Assessment Network. Equation \ref{eq:7} changes to \( \Tilde{V_{ii'}} = \sum_{j=1}^{m} \frac{1}{1+e^{-(\rho^{a_i}_j-\rho^{a_i'}_j)}} \).
    \item \textbf{w/o Preference}: The model removes individual value preferences, evaluating actions based purely on their objective scores from the Value Assessment Network. Equation \ref{eq:7} changes to \( \Tilde{V_{ii'}} = \sum_{j=1}^{m} V_{ii',j} \).
    \item \textbf{w/o Subjective}: The model removes subjective preference adjustments, using only objective action and scenario scores without individual bias adaptation. Equation \ref{eq:4} changes to \( r_j = o_j \).
    \item \textbf{w/o Scenario}: The model ranks actions solely based on their individual value impact, without incorporating scenario-based scaling in the final decision. Equation \ref{eq:5} changes to \(r_{i,j} = r_{j}^{a_i} \).
\end{enumerate}

\autoref{tab:ablation} compares the performance of different model variants, with the full DMM achieving the highest scores, demonstrating the importance of scenario-aware evaluation and subjective preference modeling for human-aligned decision-making. Removing scenario integration (\textbf{w/o Scenario}) significantly reduces performance, underscoring the necessity of context-aware decision-making, while eliminating subjective preference modeling (\textbf{w/o Subjective}) lowers accuracy, confirming the role of individual biases in aligning decisions with human choices. The \textbf{w/o Preference} and \textbf{Only Action} models perform the worst, highlighting that ignoring personalized preferences or context leads to ineffective decision-making. These results validate our two-stage framework, showing that integrating objective action impact, scenario-based adjustments, and individual preference adaptation improves decision accuracy and interpretability.

\section{Related Works}

\subsection{Decision Making with Language Models}
The evolution of language models in decision-making progresses through three distinct phases. Foundational systems such as WebGPT \citep{nakano2021webgpt} demonstrated autonomous web navigation for complex queries but relied on opaque reasoning processes and intensive human supervision—shortcomings subsequently mirrored in conversational agents including BlenderBot \citep{shuster2022blenderbot} and Sparrow \citep{glaese2022improving}. Contemporary approaches explicitly address these limitations through structured reasoning frameworks: Chain-of-Thought (CoT) prompting \citep{wei2022chain} encourages LLMs to generate step-by-step reasoning traces, while Inner Monologue \citep{huang2022inner} specializes in robotic control by integrating environmental sensorimotor feedback into its reasoning cycles. 

In parallel, research has also explored value-based approaches to guide LLM decision-making. For example, approaches in reinforcement learning from human feedback (RLHF) \citep{christiano2017deep} and related techniques like Direct Preference Optimization (DPO) \citep{rafailov2024direct} have made progress in aligning AI decision-making with human values. Despite these advances, a fundamental limitation emerges in the absence of formal value reasoning mechanisms, as current systems focus on procedural transparency while neglecting to model ethical principles, social norms, or contextual value hierarchies.

\subsection{Role-Playing Language Agents for Value Alignment}
Role-Playing Language Agents (RPLAs) leverage large language models to simulate human-like personas through contextual adaptation, enabling fine-grained value alignment beyond statistical preference aggregation \citep{chen2024persona}. Current implementations adopt three persona paradigms:
\textbf{Demographic personas} apply population-level stereotypes (e.g., occupational roles) to approximate social norms \citep{xu2023expertprompting}, while \textbf{Character personas} replicate predefined figures' behavioral patterns for value consistency \citep{wang2023rolellm}. The most adaptive \textbf{Individualized personas} dynamically evolve through user interactions \citep{salemi2023lamp}.

Recent advances in retrieval-augmented generation enhance persona grounding by integrating external knowledge, and extended memory modules support preference evolution \citep{chen2024persona}. However, most RPLAs remain confined to conversational emulation rather than implementing verifiable value-driven decision frameworks, particularly in complex action selection scenarios.

\subsection{Multi-Criteria Decision Making for Preference Modeling}
Multi-Criteria Decision Analysis (MCDA) establishes systematic frameworks for reconciling competing objectives. Foundational approaches like Multi-Attribute Utility Theory (MAUT) \citep{dyer2016multiattribute} formalize decision criteria through utility aggregation, while Analytic Hierarchy Process (AHP) \citep{vaidya2006analytic} decomposes complex decisions into hierarchical structures. These methodologies have been extended through techniques such as TOPSIS \citep{papathanasiou2018topsis} for distance-based alternative ranking and PROMETHEE \citep{brans2016promethee} for pairwise preference flow analysis, demonstrating versatility across operational research domains.

\section{Conclusion}
In this paper, we presented ValuePilot, a two-phase framework aimed at enhancing personalized decision-making in AI. The framework includes DGT for generating realistic, value-guided scenarios and DMM for ranking actions based on individual preferences. Through our experiments, DMM demonstrated a strong alignment with human decision-making, outperforming existing LLMs in replicating human choices. This research marks a preliminary step towards more personalized, value-driven AI systems. We hope it will encourage further exploration and development in this direction within the AI community.

\section{Limitations}

This study represents an initial foray into value-driven decision-making, a domain still in its formative stages within AI research. While our framework demonstrates promising alignment with human preferences, several inherent constraints merit discussion, including potential risks in scaling and generalizing across diverse applications.

As one of the earliest systematic studies in value-driven decision-making—a domain lacking established methodologies—our focus lies in validating the feasibility of aligning AI decisions with human values. While our lightweight network architecture achieves superior alignment performance compared to resource-intensive LLMs (e.g., GPT-4o, Claude 3.5 Sonnet) in our defined task space, its adaptability to highly complex scenarios involving multi-layered value conflicts remains to be explored. However, the broader question of how such architectures might scale to accommodate highly interdependent value hierarchies (e.g., societal norms interacting with personal ethics) remains an open research challenge for the field.  

While our framework's modular structure theoretically accommodates additional value dimensions, empirical validation has focused on six psychologically-grounded constructs. This intentional scope limitation enables rigorous evaluation but leaves unexamined how cultural or context-specific values might necessitate adaptations to our value quantification mechanisms. Future work must establish systematic protocols for dimension selection and cross-cultural validation.  

The reliance on synthetic data generation, while necessitated by the absence of established benchmarks for value alignment, introduces inherent constraints. Our two-stage quality assurance protocol—automated filtering, and human review—substantially mitigates but cannot eliminate potential artifacts in scenario formulation. The reliance on synthetic data generation poses a risk of overfitting to artificial scenarios, potentially limiting the model's ability to generalize to complex, and dynamic value contexts. This limitation underscores an urgent need for interdisciplinary collaborations to develop ethically-curated, demographically-diverse datasets that capture the nuanced expression of human values across contexts.  

These considerations collectively highlight the dual nature of our contributions: both as a functional framework for value-aware decision-making and as a catalyst for identifying fundamental research questions in AI value alignment. Addressing these challenges will require sustained efforts across machine learning, behavioral science, and ethics to establish rigorous theoretical foundations for computational value modeling.





\newpage
\clearpage

\appendix

\setcounter{secnumdepth}{1} 

%


\title{Technical Appendix}

\section{Experimental Dataset Considering Six Value Dimensions}\label{appendix:dataset}
\subsection{Why these six values}
In the evolving landscape of home environment multi-agent systems, understanding the underlying value dynamics that drive human interactions and choices is paramount. Our work delves into six core value dimensions—\textit{Curiosity}, \textit{Energy}, \textit{Safety}, \textit{Happiness}, \textit{Intimacy}, and \textit{Fairness}—and examines their interplay within the context of Maslow's hierarchy of needs \citep{maslow1943theory}, which provides a foundational framework for understanding human motivation and behavior.

\textbf{Broad Coverage Across Interactions:} These six value dimensions encompass both single-agent and multi-agent interactions. This dual focus ensures a comprehensive understanding of the diverse ways in which agents interact with their environments. \textit{Curiosity}, \textit{Energy}, and \textit{Safety} are mainly the value dimensions of a single agent, often involving the agent's interaction with the environment itself. In contrast, multi-agent interactions such as \textit{Intimacy}, \textit{Happiness}, and \textit{Fairness} involve complex interplays between multiple agents and humans. \textit{Intimacy} and \textit{Happiness} are deeply intertwined with social needs, while \textit{Fairness} emerges as a critical dimension in multi-agent settings, crucial for understanding human cognition and behavior \citep{sharma2022human}.

\textbf{Minimal Overlap Between Dimensions:} These six dimensions exhibit distinct characteristics, minimizing overlap. While some interactions may touch upon multiple dimensions, their unique aspects remain discernible. For instance:
\begin{itemize}
    \item Curiosity-driven exploration (\textit{Curiosity}) does not necessarily overlap with physical energy expenditure (\textit{Energy}).
    \item Ensuring a secure home environment (\textit{Safety}) differs from seeking happiness through social connections (\textit{Intimacy}) \citep{sharma2022human}.
\end{itemize}

\textbf{Alignment with Maslow's Hierarchy of Needs:} Through a meticulous selection process, these dimensions were identified as pivotal in shaping human behavior in domestic settings. We employ a multi-layered analytical approach, integrating sociological and psychological theories to substantiate the relevance of each dimension \citep{haucap2014happiness, maslow1943theory}.

\begin{itemize}
    \item \textbf{Curiosity}, drives individuals to explore and understand their environment, a manifestation of self-actualization needs that propel personal growth\citep{schwartz2012overview}.
    \item \textbf{Energy}, as a representation of physiological needs, is essential for maintaining an active and engaging lifestyle. Access to energy significantly impacts well-being and daily activities\citep{herendeen1978energy}.
    \item \textbf{Safety} is paramount in ensuring a stable and secure living space, directly impacting one's ability to thrive. A secure environment is crucial for mental and physical health, influencing overall quality of life\citep{sokolowski2022energy}.
    \item \textbf{Happiness}, while not explicitly named in Maslow's original hierarchy, can be seen as an overarching outcome that arises when the other needs are met, particularly those related to 'Esteem needs' and 'Self-actualization'\citep{forootan2022machine}.
    \item \textbf{Intimacy} fulfills the inherent need for connection and belonging, which is crucial for psychological well-being and is considered one of the fundamental aspects of human motivation\citep{mardani2015sustainable}.
    \item \textbf{Fairness} is often associated with the ‘Belongingness and Love needs' as well as the ‘Esteem needs'. This is because a sense of fairness is foundational to being accepted and respected by others. When individuals perceive fairness in social interactions, they are more likely to feel a sense of belonging and respect, which helps to fulfill the intermediate levels of Maslow's hierarchy \citep{maslow1943theory}. 

\end{itemize}


The selection of these six value dimensions—Curiosity, Energy, Safety, Happiness, Intimacy, and Fairness—intentionally encompasses the full spectrum of Maslow's hierarchy of needs. This comprehensive coverage ensures that our value dynamics analysis of agents' behaviour is grounded in a well-established psychological framework, providing a robust basis for understanding the multifaceted nature of these interactions within home environments. These six dimensions are more frequently encountered in home environments as they directly relate to daily living and personal fulfillment. Other values from Maslow's theory, such as physiological needs beyond basic energy requirements, tend to manifest less frequently in the context of routine home life. 

Based on our model's considerations, we have selected these six value dimensions as focal points for our analysis, as they offer a pertinent and practical lens through which to examine the prevalent dynamics in human-agent interactions and strive to enhance the symbiotic relationship between humans and agents.

\subsection{Statistics}\label{appendix:dataset_statistics}
\begin{figure*}[ht]
    \centering
    \begin{subfigure}[b]{0.28\textwidth}
        \centering
        \includegraphics[width=\textwidth]{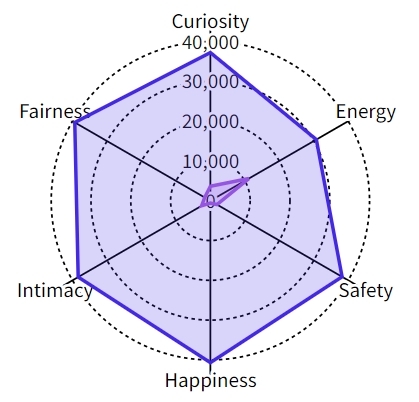}
        \subcaption{Value Distribution in Trainset}
        \label{fig:image1}
    \end{subfigure}
    \hfill
    \begin{subfigure}[b]{0.28\textwidth}
        \centering
        \includegraphics[width=\textwidth]{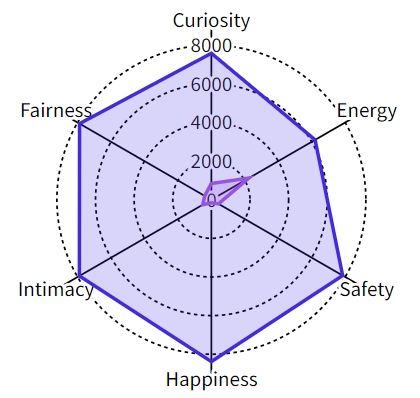}
        \subcaption{Value Distribution in Testset}
        \label{fig:image3}
    \end{subfigure}
    \hfill
    \begin{subfigure}[b]{0.28\textwidth}
        \centering
        \includegraphics[width=\textwidth]{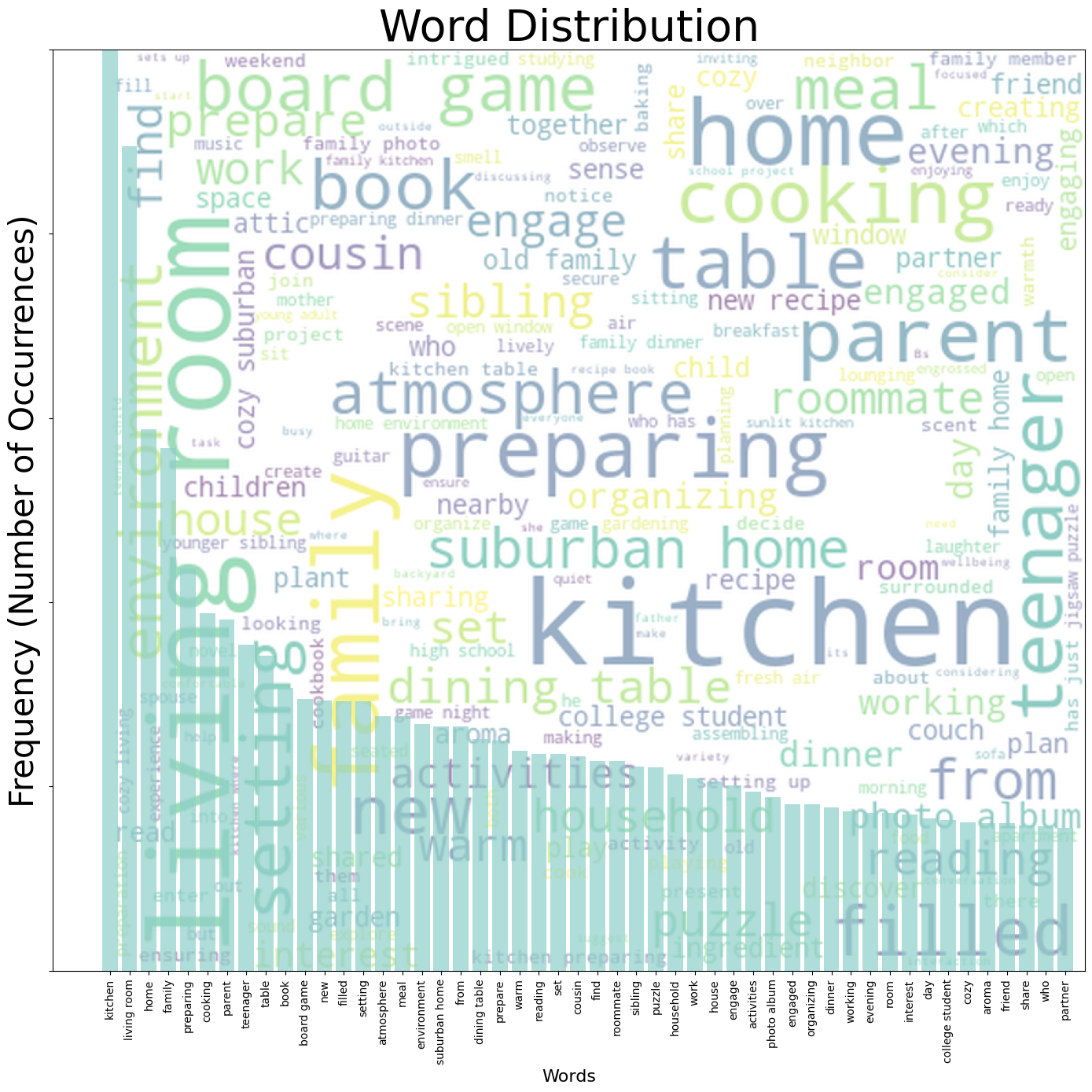}
        \subcaption{Word Frequency Distribution}
        \label{fig:Word Frequency Distribution}
    \end{subfigure}
    \hfill
    \begin{subfigure}[b]{0.28\textwidth}
        \centering
        \includegraphics[width=\textwidth]{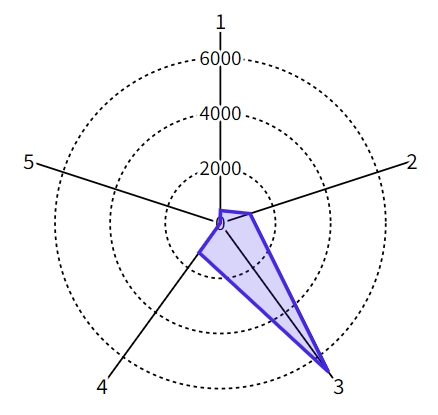}
        \subcaption{Agent Number in Trainset}
        \label{fig:Agent Number in Trainset}
    \end{subfigure}
    \hfill
    \begin{subfigure}[b]{0.28\textwidth}
        \centering
        \includegraphics[width=\textwidth]{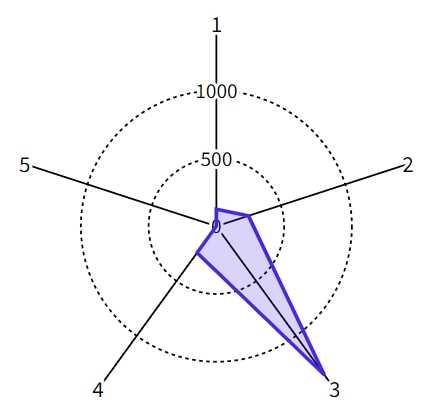}
        \subcaption{Agent Number in Testset}
        \label{fig:Agent Number in Testset}
    \end{subfigure}
    \hfill
    \begin{subfigure}[b]{0.28\textwidth}
        \centering
        \includegraphics[width=\textwidth]{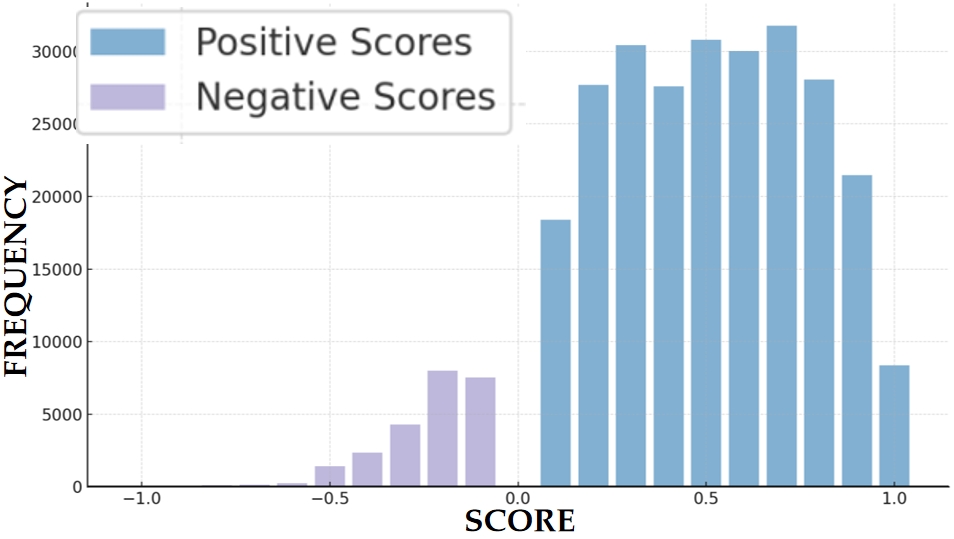}
        \subcaption{Score Frequency Distribution}
        \label{fig:image4}
    \end{subfigure}
    \caption{Statistical characteristics of the dataset; (a) and (b) show the number of times that the scores of all six value dimensions in the dataset appeared as positive and negative values in the scenario; (c) shows the word cloud image of the words that appear in the scenerio; (d) and (e) show the number of agents in the scenarios of the dataset; (f) shows the score distribution of all value dimensions.}
    \label{fig:six_images}
\end{figure*}

For the dataset we generated, we analyzed the characteristics of the text from multiple perspectives (see \autoref{fig:six_images}).

\textbf{Hierarchical Dataset Architecture and Quality Assurance}

The proposed dataset employs a six-tier hierarchical structure (Datasets 1--6) to systematically evaluate computational models across escalating complexity levels of multi-dimensional value interactions. Each tier corresponds to distinct combinatorial configurations of six core value dimensions: \textit{curiosity}, \textit{energy}, \textit{safety}, \textit{happiness}, \textit{intimacy}, and \textit{fairness}. As detailed in \autoref{tab:dataset_stats}, Dataset 1 isolates single-value dynamics (1-D), while subsequent tiers progressively integrate dimensions up to Dataset 6, which encapsulates full six-dimensional (6-D) interactions.

\begin{table*}[htbp]
  \centering
  \caption{Hierarchical Dataset Composition with Filtering Metrics}
  \label{tab:dataset_stats}
  \begin{tabular}{cccccccccc}
    \toprule
    \textbf{Dataset} & 
    \multicolumn{4}{c}{\textbf{Data Filtering}} & 
    \multicolumn{2}{c}{\textbf{Final Scenarios}} & 
    \multicolumn{2}{c}{\textbf{Final Actions}} \\
    \cmidrule(lr){2-5} \cmidrule(lr){6-7} \cmidrule(lr){8-9}
    & 
    \multicolumn{2}{c}{Train} & 
    \multicolumn{2}{c}{Test} & 
    Train & Test & Train & Test \\
    \cmidrule(lr){2-3} \cmidrule(lr){4-5} \cmidrule(lr){6-7} \cmidrule(lr){8-9}
    & 
    Removed & Rate (\%) & 
    Removed & Rate (\%) & 
    & & & \\
    \midrule
    1-D & 2,333 & 12.98 & 430 & 12.08 & 1,771 & 355 & 15,644 & 3,129 \\
    2-D & 2,632 & 15.14 & 430 & 12.25 & 1,689 & 353 & 14,753 & 3,080 \\
    3-D & 3,264 & 18.16 & 501 & 17.20 & 1,704 & 342 & 14,712 & 2,411 \\
    4-D & 3,688 & 22.36 & 728 & 20.59 & 1,609 & 332 & 12,815 & 2,807 \\
    5-D & 3,584 & 23.36 & 863 & 24.38 & 1,412 & 323 & 11,761 & 2,677 \\
    6-D & 4,083 & 22.85 & 901 & 25.17 & 1,714 & 334 & 13,788 & 2,678 \\
    \midrule
    \textbf{Total} & 19,584 & 19.00 & 3,853 & 18.67 & 9,899 & 2,039 & 83,473 & 16,782 \\
    \bottomrule
  \end{tabular}
  \vspace{0.2cm}
\end{table*}

To mitigate potential biases in generated content, we implement a two-stage quality assurance protocol. First, an automated procedure removes value-related pairs explicitly mentioned in input prompts from both scenarios and actions. As quantified in \autoref{tab:dataset_stats}, this eliminates 12.98\%--23.36\% of training actions and 12.08\%--25.17\% of test actions across complexity tiers. Second, the manual phase involves iterative refinement to correct implausible scenarios and inconsistent actions, prioritizing behavioral coherence, interaction plausibility, and dimensional proportionality.

Aggregate statistics confirm the dataset's stratified design, containing 9,899 training scenarios (83,473 actions) and 2,039 test scenarios (16,782 actions) across all tiers. This architecture enables granular analysis of model robustness against dimensional scalability while maintaining methodological rigor through systematic bias mitigation.

\textbf{Distribution of six-value dimensions (\autoref{fig:image1} and \autoref{fig:image3}): }We counted the number of times that the scores of all six value dimensions in the dataset appeared as positive and negative values in the scenario. Most of the scores of the six value dimensions appeared as positive values, and the total number of the six dimensions is basically the same, though the negative scores of \textit{Energy} are relatively more. The blue box line represents a positive value, and the purple box line represents a negative value. 

\textbf{Word Frequency Analysis (\autoref{fig:Word Frequency Distribution}): }We performed word frequency analysis on the dataset scenes after removing \textit{stop words}, which are items with less statistical significance such as prepositions, adverbs, quantifiers, etc., and found many interesting phenomena. Among the nouns, \textit{kitchen} and \textit{living room} appear most frequently, representing the places where scenarios often occur. \textit{Preparing} and \textit{cooking} are the most common verbs, representing actions that often occur in the scenario. Interestingly, \textit{board game} also appears frequently in our dataset. We think that it is reasonable to play board games as a way of multi-person interaction in a home environment that is compatible with happiness, intimacy, and fairness at the same time.

\textbf{Number of Agents (\autoref{fig:Agent Number in Trainset} and \autoref{fig:Agent Number in Testset}): }We found that the number of agents in the scenarios of the dataset ranges from 1 to 5, and the number of scenarios containing 3 agents is the largest. In the training set and the test set, the corresponding proportion of the number of agents is basically the same.

\textbf{Score Frequency Distribution (\autoref{fig:image4}): }We calculated the score distribution of all value dimensions and found that values from -1.0 to +1.0 are represented. The score distribution between -0.1 and +1.0, which we pay more attention to, is relatively uniform.

\section{Action Selection Module}\label{appendix:action_selection}
Algorithm \ref{alg:action-selection} details the procedure used by the Action Selection Module to integrate the Value Assessment Network's scoring output with pre-personalized preference values, resulting in a ranked sequence of actions.

\section{Model Training and Experiment}\label{appendix:model}

\subsection{Implementation Details}
In Experiment 1, we train the networks on a train set comprising 80\% of the total dataset, and then test it on 20\% of the total dataset, aiming to show the efficiency of Value Assessment Network. The experiments are conducted on a system with a single NVIDIA GeForce RTX 4090 GPU (24 GB memory), using CUDA Version 12.6 and driver version 560.70. The models were optimised using the Adam optimizer \citep{kingma2014adam}. The hyperparameters in this optimal configuration consisted of a learning rate range of [1e-3, 1e-5] and a weight decay range of [1e-5, 1e-4] for the Value Assessment Network during Phases 1 and 2. The total training time was approximately 15 minutes.

\subsection{Model Training Procedure}
We use a curriculum learning approach \citep{bengio2009curriculum} to train our models, helping our model build a more robust understanding of basic concepts before tackling more challenging scenarios. To ensure balanced representation, the entire dataset is divided into \( l \) (4 in our experiment) chunks. Each chunk contains a mix of data generated from all combinations of \( n \) (6 in our experiment) value dimensions. The training process follows a predefined order of these groups. This sequence is designed to help the model to adapt and learn efficiently, minimizing the risk of overfitting or underfitting.

Algorithm \ref{alg:training} provides a detailed overview of the curriculum learning approach used to train our models, guiding them through progressively complex data to enhance learning efficiency and performance. In the context of the algorithm you provided, the sequence of dataset $S_{data}$ refers to the predefined order in which different subsets of your data are used during the model training process. This sequence is crucial in the curriculum learning approach because it dictates the order in which the model encounters data of varying difficulty or complexity.

\begin{algorithm}[H]
\caption{Model Training Procedure}
\label{alg:training}
\begin{algorithmic}[1]
\STATE \textbf{Set the sequence of dataset:} $S_{data}= $ [6,1,5,2,4,3]
\STATE \textbf{Initialize} number of chunks to \( l \)

\FOR{each chunk $i$ from 0 to \( l \) - 1}
    \STATE Initialize Datasets
    
    \FOR{each \texttt{n} in training order}
        \STATE \texttt{Combine\_Dataset}.\texttt{concat}(\texttt{Dataset}[\texttt{n}])
        \STATE Initialize early stopping parameters
        
        \FOR{each epoch}
            \STATE Train model, validate model, log metrics, and check for early stopping
            \IF{early stopping criteria are met}
                \STATE Save model
            \ENDIF
        \ENDFOR
    \ENDFOR
\ENDFOR

\end{algorithmic}
\end{algorithm}

\subsection{Additional Results}
\subsubsection{Hyperparameters}
As shown in \autoref{tab:training_results}, Our Networks are trained with different training order and hyperparameters. The results show that a training order of [6, 1, 5, 2, 4, 3] and [lr, wd] sets of {[1e-3, 1e-5], [1e-5, 1e-4]} reaches the highest accuracy and the shortest training time.

\begin{table}
\centering
\setlength{\tabcolsep}{3pt}
\renewcommand{\arraystretch}{0.9}
\small
\begin{tabular}{c>{\centering\arraybackslash}p{0.65cm}>{\centering\arraybackslash}p{0.65cm}>{\centering\arraybackslash}p{0.65cm}>{\centering\arraybackslash}p{0.65cm}|>{\centering\arraybackslash}p{1.2cm}p{1cm}}
\toprule
\multirow{1}{*}{\textbf{Training}} & \multicolumn{2}{c@{}}{\textbf{Phase 1}} & \multicolumn{2}{c@{}}{\textbf{Phase 2}} & \textbf{Time} & \textbf{Avg} \\
\textbf{Order} & {lr} & {wd} & {lr} & {wd} & \textbf{(min:sec)} & \textbf{Acc} \\ \midrule
3,4,2,5,1,6 & 1e-4 & 1e-5 & 1e-5 & 1e-4 & 22:43 & 66.06\% \\ 
6,1,5,2,4,3 & 1e-4 & 1e-5 & 1e-5 & 1e-4 & 23:25 & 66.15\% \\
1,2,3,4,5,6 & 1e-4 & 1e-5 & 1e-5 & 1e-4 & 22:23 & 66.00\% \\ 
6,5,4,3,2,1 & 1e-4 & 1e-5 & 1e-5 & 1e-4 & 24:53 & 63.80\% \\ 
6,1,5,2,4,3 & 1e-4 & 1e-5 & 1e-4 & 1e-4 & 23:25 & 65.00\% \\ 
\textbf{6,1,5,2,4,3} & \textbf{1e-3} & \textbf{1e-5} & \textbf{1e-5} & \textbf{1e-4} & \textbf{14:16} & \textbf{66.70\%} \\
6,1,5,2,4,3 & 1e-5 & 1e-5 & 1e-5 & 1e-4 & 133:50 & 66.06\% \\
\bottomrule
\end{tabular}
\caption{Comparison of Different Training Orders and Hyperparameter Configurations. \textbf{lr} stands for learning rate. \textbf{wd} stands for weight decay. Threshold $t = 0.2$ for AvgAcc.}
\label{tab:training_results}
\vspace{-6pt}
\end{table}

\subsubsection{Encoder Model}
To prove that our decision making network works well for a variety of relatively small language models, We replaced the encoder's model, and the experimental results are shown in \autoref{tab:Encoder_selection}.

\begin{table}
    \centering
    \begin{tabular}{cc}
    \toprule 
    \textbf{Encoder}& \textbf{Average Accuracy (\%)}\\ 
    \midrule
    \textbf{T5-base}& \textbf{66.70}\\ 
    Flan-T5&  66.00\\ 
    BERT&  65.45\\ 
    RoBERTa&  65.69\\
    \bottomrule
    \end{tabular}
    \caption{Selection study of encoder models. Threshold $t = 0.2$ for Average Accuracy.}
    \label{tab:Encoder_selection}
\end{table}

The T5-base model outputs the highest Average Accuracy. However, upon transforming different encoder models, the final similarity remains relatively consistent, which proves that our fine-tuning method is applicable to different pre-trained language models.

\section{Human Study}\label{appendix:human_evaluation}
\subsection{Demographic Analysis}\label{appendix:Demographic}

The study cohort (N=40) exhibited substantial demographic diversity across age, ethnicity, and socioeconomic dimensions. Subjects spanned adolescents (12.5\%, n=5), young adults (47.5\% aged 18-25), and middle-aged individuals (2.5\% aged 51-60), with balanced gender representation (55.0\% female, 45.0\% male). Educational attainment ranged from secondary education to advanced degrees, with 75\% holding bachelor's qualifications or higher. Occupational diversity spanned 11 distinct professional categories, including academic researchers (15.0\%), technical specialists (7.5\%), and commercial professionals (5.0\%), though students predominated (52.5\%). While the sample skews toward younger, educated urban populations—a common limitation in survey-based studies—this demographic orientation does not substantially compromise the internal validity of findings, though broader generalizability to rural or less-educated groups requires confirmation through targeted sampling in future studies.

\begin{figure*}[ht]
    \centering
    \includegraphics[width=0.95\textwidth]{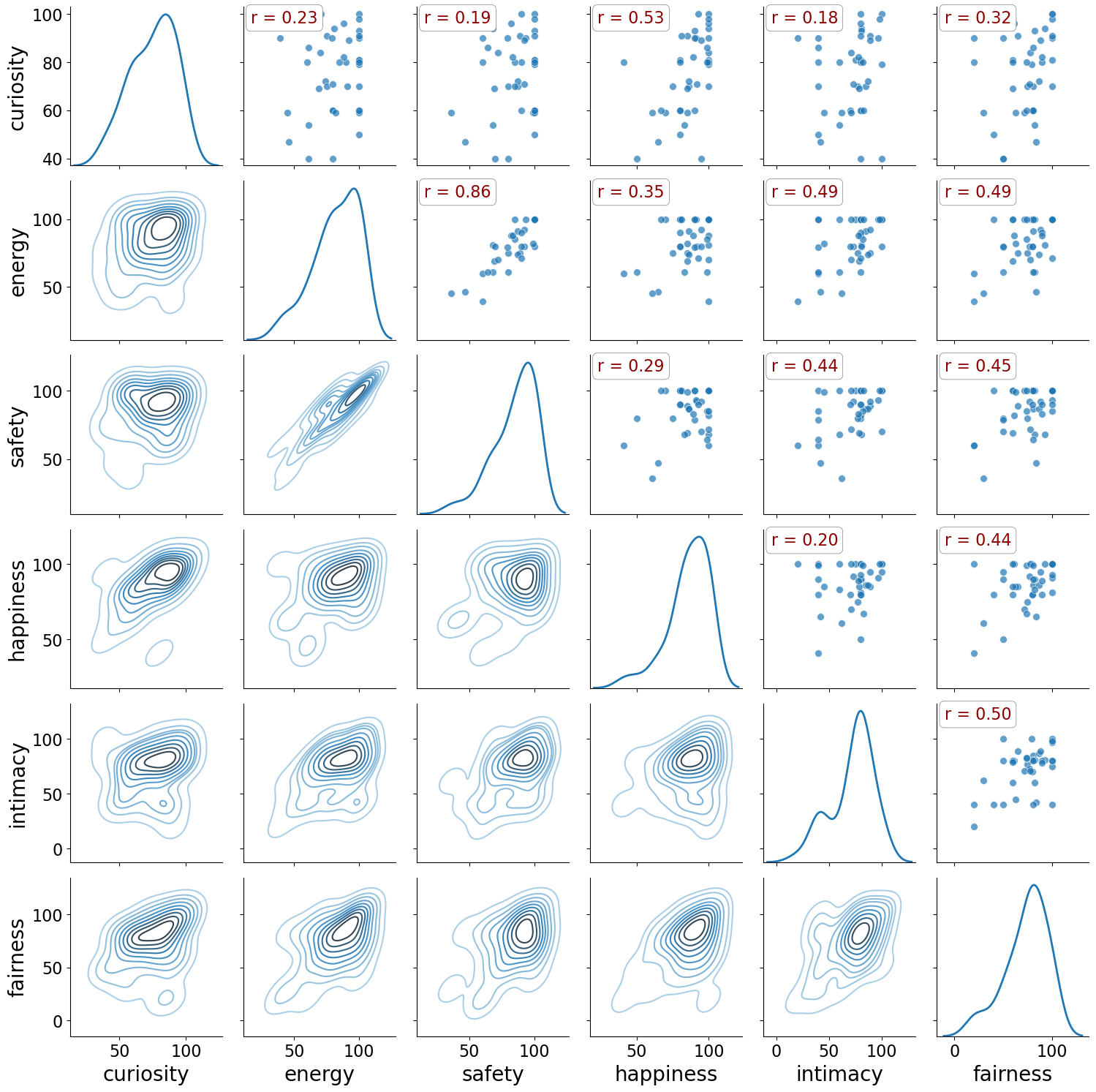}
    \caption{Pairwise Correlations and Distributions of Value Preferences Across the Six Dimensions with Pearson correlation coefficient (r)}
    \label{fig:prefcorr}
\end{figure*}

\begin{figure}[ht]
    \centering
    \includegraphics[width=0.95\columnwidth, keepaspectratio]{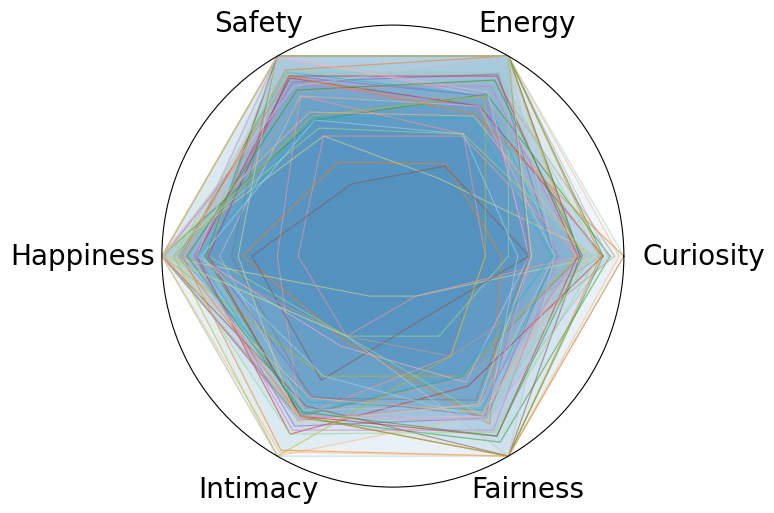}
    \caption{Radar chart showing the value preferences of all subjects. Each polygon represents a subject, and different colors of outline correspond to different subjects.}
    \label{fig:radar}
\end{figure}

In addition to these demographic characteristics, we collected subjects' preference ratings on six value dimensions: \emph{curiosity}, \emph{energy}, \emph{safety}, \emph{happiness}, \emph{intimacy}, and \emph{fairness}. As shown in \autoref{fig:prefcorr} and \autoref{fig:radar}, these preference distributions were notably broad, indicating that our sample captured a wide range of individual differences in value orientations. The pairwise correlation coefficients among the dimensions are also presented, revealing varying degrees of interrelationship across curiosity, energy, safety, happiness, intimacy, and fairness.

\begin{figure}[ht]
    \centering
    \includegraphics[width=0.95\columnwidth,
                    keepaspectratio]{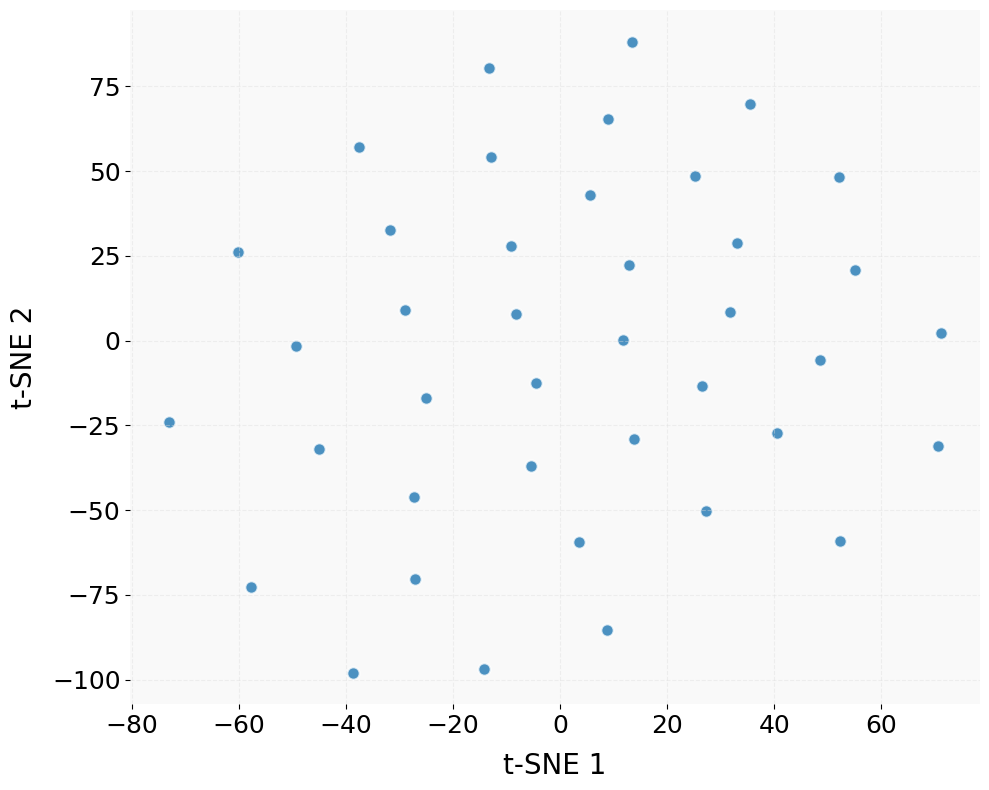}
    \caption{t-SNE Projection of All Subjects' Value Preferences}
    \label{fig:pref_tsne}
\end{figure}

Furthermore, \autoref{fig:pref_tsne} provides a two-dimensional t-SNE visualization of the subjects' preference profiles. The relatively dispersed clustering in this plot suggests that there is no single dominant pattern; instead, subjects' preferences exhibit substantial heterogeneity. Taken together, the breadth of preference distributions, alongside the demographic diversity described earlier, underscores the representativeness of the sample in terms of varying backgrounds and value priorities.

\subsection{Comparative Experiment}
We designed a test questionnaire containing 11 formal test questions and collected data from 40 subjects. The questionnaire included a detailed introduction to six value dimensions at the beginning, where subjects rated the importance of these dimensions in their daily lives on a scale from 0 to 1, with higher scores indicating greater importance.

To ensure subjects fully understood the experiment's design, a simulation test with three preliminary questions was conducted. This allowed subjects to familiarize themselves with the process, enabling them to provide more accurate data during the formal test. Following the simulation test, subjects re-scored the importance of the six value dimensions, generating a final scoring vector. For each scenario described in the questionnaire, subjects were asked to rank their willingness to choose different actions based on their value dimension scores. This produced a six-dimensional value scoring vector and corresponding action ranking data. Subjects also identified the value dimensions associated with each action in the scenarios, which facilitated the analysis of the rationality of the data. Based on this analysis, we screened and reorganized the questionnaires, resulting in data from 10 subjects.

We then input the scenario descriptions and corresponding action lists into our model, along with each subject’s six-dimensional value scoring vector. The model processed this data and outputted action selection rankings based on the subject's value vector. As a comparative experiment, we also input the six-dimensional scores of the subjects into large language models (Claude-3.5-sonnet \citep{anthropic2024claude}, Llama-3.1-70b \citep{dubey2024llama}, Llama-3.1-405b \citep{dubey2024llama}, Gemini-1.5-pro \citep{team2023gemini}, Gemini-2-flash \citep{team2023gemini}, GPT-4o \citep{achiam2023gpt} and GPT-4o-mini \citep{achiam2023gpt}) via input prompts. These models provided action selection rankings based on the scores and scenario descriptions.

For each participant, we obtained action selection rankings from both our model and the large language models. We then used the proposed sequence similarity measurement method OS-Sim and First-Acc to compare the ranking lists generated by our model and the large language models with the actual lists provided by the subjects.

Then we calculate the Mean OS-Sim, based on the comparative results of different models, fitting the choice of human subjects. Through confidence analysis, we found that the fitting degree of our model to human action selection is better than that of the current mainstream large language model, which proves the effectiveness of our model in analyzing the value dynamics of agents in the home environment.

We also conducted a experiment to compare the effectiveness of different Multi-Criteria Decision Analysis (MCDA) methods in accurately ranking actions within a given scenario. Based on the experimental results, PROMETHEE shows the highest average accuracy at 73.16\% with a low error margin of ±0.43\%, making it the most reliable method in this scenario, as shown in \autoref{tab:MCDA}.

\begin{table}
\centering
\setlength{\tabcolsep}{4pt}
\begin{tabular}{cc}
\toprule
\textbf{Method} & \textbf{Mean OS-Sim (\%)} \\
\midrule
\textbf{PROMETHEE} & \textbf{73.16 $\pm$ 0.43} \\
AHP & 72.9 $\pm$ 0.56 \\
MAUT & 67.08 $\pm$ 0.47 \\
TOPSIS & 68.19 $\pm$ 0.53 \\
\bottomrule
\end{tabular}
\caption{Comparison of Different Ranking Methods}\label{tab:MCDA}
\vspace{-6pt}
\end{table}

\subsection{Test Questionnaire For Experimenters}
\textbf{Test Instructions:}
We have constructed a value-driven decision-making model for complex environments. The input to the model consists of six value dimensions:
\begin{itemize}
    \item \textbf{Curiosity:}
    An intrinsic desire for knowledge about the world around us. It motivates individuals to actively seek new information and experiences, expanding their understanding of things and enriching their perspectives through exploration. People with high curiosity tend to: actively seek knowledge, bravely explore unknown areas, ask questions to deepen understanding, maintain an open mind, and accept new and diverse views.
    \item \textbf{Energy:}
    The pursuit and satisfaction of physiological energy needs, including food, water, air, sleep, and exercise. It means striving for enough energy to maintain physical and mental health. People with high energy needs tend to: maintain a healthy diet, ensure adequate hydration, secure enough and quality sleep, engage in moderate exercise, and breathe fresh air regularly.
    \item \textbf{Safety:}
    The concern for one's physical and psychological health and the need for stability and predictability in one's environment. This value dimension triggers actions to ensure personal safety, stability in life, and protection from pain, threats, and illnesses, as well as safeguarding personal property. People with high safety needs tend to: maintain physical safety measures, cultivate and maintain mental health, seek stable living conditions, plan for emergencies, and protect personal property.
    \item \textbf{Happiness:}
    A desire for positive emotions and a sense of fulfillment. It drives actions that bring joy, including cultivating positive relationships, achieving personal goals, and enjoying pleasant activities. Happiness manifests in the real world as pleasure and satisfaction with life. People with high happiness tend to: pursue positive emotions, establish meaningful relationships, strive towards personal goals, seek enjoyment and entertainment, and cultivate a sense of accomplishment.
    \item \textbf{Intimacy:}
    Emphasizes deep connections and emotional resonance in various relationships, including family, friendship, romantic, and subordinate relationships. It reflects actions taken to achieve mutual support, understanding, and shared experiences. In the real world, intimate relationships provide warmth and support, promoting individual psychological health and well-being. People with high needs for intimacy tend to: share thoughts, feelings, and experiences, provide support and understanding, willingly share personal privacy, engage in positive and meaningful communication, and pursue a sense of belonging.
    \item \textbf{Fairness:}
    Reflects an individual's pursuit of justice and equality. It motivates actions to create fair and equal conditions in the real world, including actions to ensure fair treatment in social and organizational environments. The goal is to ensure that every individual has equal opportunities and rights. People who highly value fairness tend to: respect equal rights and treatment, promote fair opportunities for everyone, participate in promoting equal decision-making, and treat others justly and fairly.
\end{itemize}
We hope you fill in the questionnaire. We ensure that the questionnaire is filled anonymously and the obtained data is only used to test the generalization effect of the model in the home environment without commercial use.

\textbf{Simulation Test:}
Before addressing specific questions, there is a simulation test to help you understand the specific format of the test.

Please rate your attention to the six value dimensions in your daily life. Ratings are on a scale from 0 to 1, with higher scores indicating greater attention to the value dimension.
\begin{itemize}
    \item \textbf{Curiosity:}\_\_\_\_\_\_\_
    \item \textbf{Energy:}\_\_\_\_\_\_\_
    \item \textbf{Safety:}\_\_\_\_\_\_\_
    \item \textbf{Happiness:}\_\_\_\_\_\_\_
    \item \textbf{Intimacy:}\_\_\_\_\_\_\_
    \item \textbf{Fairness:}\_\_\_\_\_\_\_
\end{itemize}

Below are three simulation test questions below. Please bring the first-person perspective into the following scene. What actions do you tend to choose in this scene?

Please fill in the form below each question. 
The first step, for each action, please give what you think the value dimension of the action is. 
\textbf{For each action, please indicate the value dimensions it represents by placing a checkmark (\checkmark) in the appropriate boxes.}

\vspace{1em} 

\begin{minipage}{0.4\textwidth}
    \resizebox{\textwidth}{!}{
        \begin{tabular}{|c|c|c|c|c|}
            \hline
            Action & Curiosity & Energy & Safety & Happiness \\ \hline
            1 & \checkmark &  &  & \checkmark \\ \hline
            2 &  &  &  & \checkmark \\ \hline
            3 &  & \checkmark &  & \checkmark \\ \hline
            ... & & & & \\ \hline
        \end{tabular}
    }
\end{minipage}

\vspace{1em} 

\begin{minipage}{0.4\textwidth}
    \resizebox{\textwidth}{!}{
        \begin{tabular}{|c|c|c|c|}
            \hline
            Action & Intimacy & Fairness & Other Values\\ \hline
            1 &  &  &  \\ \hline
            2 & \checkmark &  &  \\ \hline
            3 &  &  &  \\ \hline
            ...& & & \\ \hline
        \end{tabular}
    }
\end{minipage}

\vspace{1em} 

\textbf{Examples of other value dimensions:} \\
1 Responsibility: Taking the initiative to pick up waste paper falling on the ground and throw it into the trash can may be a test of responsibility.\\
2 Ambitiousness: Choosing to read at home on a good weekend may be out of ambition. \\
3 Authority: Forcing children to do something may be out of the consideration of maintaining their own family authority. 
Etc.

The second step is to prioritize each action in the ' action selection ' according to your willingness to choose, or the possibility of you making the action from high to low. (Blank box to fill in each action before the corresponding serial number 1234...) 
(Note: Different actions are selected for the parallel occurrence relationship in the same scene, and do not arrange in the order of the occurrence time in the scene.)

\begin{center}
\begin{tabular}{|c|c|c|c|c|c|}
    \hline
    Action Preference Sort & 1 & 2 & 3 & 4 & ...\\ \hline
    Action Index &  &  &  &  &  \\ \hline
\end{tabular}
\end{center}

\textbf{Formal Test:}
Presumably, you already know how to test. Here is a formal test with 11 questions.

Please re-score yourself on the importance of the six value dimensions in daily life. Ratings are on a scale from 0 to 1, with higher scores indicating greater attention to the value dimension.

\begin{itemize}
    \item \textbf{Curiosity:}\_\_\_\_\_\_\_
    \item \textbf{Energy:}\_\_\_\_\_\_\_
    \item \textbf{Safety:}\_\_\_\_\_\_\_
    \item \textbf{Happiness:}\_\_\_\_\_\_\_
    \item \textbf{Intimacy:}\_\_\_\_\_\_\_
    \item \textbf{Fairness:}\_\_\_\_\_\_\_
\end{itemize}

Below are 11 formal questions, please finish them in the same requirements as before.

\textbf{Q1:}\\
\textbf{Scenario:} You are a 20-year-old college student. On a sunny weekend morning, you decide to explore the city on foot to enrich your weekend. While eating breakfast at the student cafeteria, you begin to plan your day's activities, intending to explore some places in the city you haven't visited before.

\textbf{Action list:}\\
1. You plan to first go to an online library to collect e-books about the city's development history, so you can have a basic understanding of the city before exploring.\\
2. You are hesitant about exploring unknown corners of the city and think it is better to stick to visiting main and officially recommended attractions to avoid unnecessary trouble.\\
3. You plan to try a mobile app to experience the city's new AR historical guide application, exploring the city in a different way through virtual reality technology.\\
4. You believe that the true joy of hiking adventures lies in discovering the unknown. So, you decide not to make any detailed plans but instead choose a small alley or street you have never walked before, following your intuition to explore the city..
\vspace{0.1cm}

\textbf{Q2:}\\
\textbf{Scenario:} You, B, and C are university roommates. The dorm is usually messy, and no one cleans the common space. Last week, during the routine dormitory cleanliness inspection, you were warned by the dormitory administrator. At this moment, the three of you are seriously discussing how to distribute household chores in the future, with a whiteboard at the door displaying the categories of chores you just listed.

\textbf{Action list:}\\
1. To prevent 'free-riding', you seriously propose to establish a detailed schedule for household chores and assign tasks to each person.\\
2. To prevent omissions, you suggest that everyone think about whether the table listed on the whiteboard fully includes all the chores.\\
3. You think it is unnecessary to assign chores too specifically since you are roommates living together, so you suggest that everyone just do their chores as they go, reminding each other when someone has time.\\
4. You don't want to make things too complicated and propose to decide the assignment of chores by rolling dice.\\
5. You suggest that it's not a big deal and not to be too serious; proposing to go out for dinner together while discussing the distribution of chores.
\vspace{0.1cm}

\textbf{Q3:}\\
\textbf{Scenario:} You, A, and B are a family living together in a suburban area of a quiet town. On the weekend, the three of you are relaxing at home, listening to soft music, enjoying the tranquility of the afternoon. The sun is shining outside, leaves gently swaying, a breeze passing by, creating a relaxed and pleasant atmosphere. You are sitting on the living room sofa, exchanging your moods and recent trivialities.

\textbf{Action list:}\\
1. To enhance family intimacy, you suggest playing a simple riddle game together.\\
2. You feel this weekend should be about relaxation, so you suggest preparing some snacks and watching a light-hearted comedy movie together.\\
3. You think it's a good opportunity to do some housework during the weekend rest time, such as cleaning the living room or organizing clutter, to make the home more tidy and comfortable.\\
4. You suddenly remember a fresh activity and suggest going for a walk in the nearby park together, breathing fresh air and relaxing.\\
5. You suggest using this time to prepare dinner together, trying out some new dishes to enhance family bonding.
\vspace{0.1cm}

\textbf{Q4:}\\
\textbf{Scenario:} You are in the kitchen adjusting the stove temperature, as dinner for your family is being cooked. B is your husband, tired from work, sitting idly in the kitchen. C is your daughter, focused on a complex school project she has been working on for a while. A severe storm has started outside.

\textbf{Action list:}\\
1. To help B relax, you go to talk to him and prepare a hot drink for him.\\
2. To prevent accidents from the storm, you check all doors and windows to ensure they are securely locked.\\
3. To prevent your daughter from being disturbed by the noise of the stove, you close the kitchen door.\\
4. To provide some suggestions and help, you discuss the project with C.\\
5. To create a cozy atmosphere for dinner, you start setting the table.\\
6. Seeing C focused, you remind her to take a break and relax, joining you for dinner later.
\vspace{0.1cm}

\textbf{Q5:}\\
\textbf{Scenario:} You are an enthusiastic middle school student interested in astronomy. B is your brother, a Ph.D. student, and C is your younger sister. It's holiday time, and all three of you are at home. You are casually lying on the living room sofa, flipping through an atlas; B is reading about aerodynamics, and C is preparing a smoothie. An unopened world map puzzle you just bought is on the table, and a computer next to the table displays a news article about breakthroughs in the field of astronomy discussing the possibility of life on an exoplanet.

\textbf{Action list:}\\
1. Looking at the world map on the table, you can't resist opening it and starting to puzzle.\\
2. You propose to C to play a quiz game about the capitals of countries using the atlas, with B acting as the judge.\\
3. Inspired by the article on the computer, you discuss with C whether you believe in the existence of extraterrestrial life.\\
4. On a whim, you build a fort of chairs and blankets on the sofa, imagining it as a spacecraft headed to the planet.
\vspace{0.1cm}

\textbf{Q6:}\\
\textbf{Scenario:} You are a high school student who has just moved to a new community. After school, you sit in your room feeling a bit bored. B, your younger brother, built a sophisticated treehouse with the help of C this morning. He is now happily playing with toys inside the treehouse, which is built very high, taller than your height. C, your neighbor, is a retired carpenter, sitting on his balcony drinking coffee, peacefully watching B play.

\textbf{Action list:}\\
1. To ensure the treehouse is sturdy and your brother is safe, you pick up a toolbox to check for any loose bolts and observe the branches to see if they are strong enough to support the treehouse.\\
2. To alleviate your boredom, you decide to bring some toys and join your brother in the treehouse to play together.\\
3. Curious about the interior structure of the treehouse, you decide to climb up and see how it is built.\\
4. Seeing the exquisitely built treehouse, you become curious about C's career and start a conversation with him.
\vspace{0.1cm}

\textbf{Q7:}\\
\textbf{Scenario:} You and B are college students living together off-campus. One hot summer noon on a weekend, both of you are at home. You go to the kitchen to find something cool to drink, open the fridge, and pour the last bit of orange juice into your cup, eager to drink it to relieve the heat. Just as you are about to finish it, you realize it's the last bottle in the fridge and notice B looking at you.\

\textbf{Action list:}\\
1. You get another cup and pour half the juice for B.\\
2. You plan to pretend you didn't notice it was the last cup, drink it all now, and apologize verbally to B later.\\
3. You plan to drink it all first, then as an apology, go out in the heat later to buy a few more bottles.\\
4. You think it's just a cup of juice and not a big deal, so you drink it all and then go back to your activities.\\
5. You suggest playing rock-paper-scissors with B, where the winner gets the juice, to make the situation fair.\\
6. You stop for a moment to think, then ask B if he wants to drink, and give the juice to him.
\vspace{0.1cm}

\textbf{Q8:}\\
\textbf{Scenario:} You are a teenage boy living in a villa, currently using 3D modeling software in your room to create a digital model of a dinosaur skeleton for a school extracurricular research project, currently facing a bottleneck and unsure how to proceed. B is your father, a paleontologist, sitting at a desk across the room, focused on researching the latest discoveries in his field. C is your pet, a gentle Golden Retriever, comfortably curled up at B's feet on a cozy rug. Near the window in the room, workers are carrying in your family's newly purchased astronomical telescope, getting ready to set it up.

\textbf{Action list:}\\
1. Attracted by the telescope, you go over to join in the excitement, curiously asking the workers about how to operate the telescope.\\
2. Deciding to shift your focus, you go to pet C, covering him with a blanket to keep warm.\\
3. You decide to talk to B about the bottleneck in making the dinosaur model, hoping B can help open your mind.\\
4. Deciding to shift your focus, you take C out to play fetch, hoping to breathe some fresh air and exercise.\\
5. You decide to take a short nap, thinking you might just be too tired, and that waking up might bring new ideas.\\
6. Deciding to shift your focus, you walk around the large house, checking to ensure all windows and doors are secure.
\vspace{0.1cm}

\textbf{Q9:}\\
\textbf{Scenario:} In a cozy family setting on a typical Thursday evening, you, B, and C, three members of the family, are discussing your weekend plans together. You really want to go see a movie, but B and C plan to do DIY cooking and play indoor video games, respectively. You hope to negotiate a consensus to improve everyone's satisfaction with the plan.

\textbf{Action list:}\\
1. You plan to create a schedule for the weekend to see if there's enough time for all activities.\\
2. You insist on going to see the movie and try to convince B and C by explaining how worthwhile the movie is.\\
3. For fairness, you suggest that everyone vote on what to do over the weekend, each person voting for one other activity besides their own.\\
4. To accommodate everyone's ideas, you start thinking about a suitable place that might allow for watching movies, cooking, and playing video games at the same time.\\
5. You suggest that everyone go to their preferred places individually, without the entire family having to stick together.
\vspace{0.1cm}

\textbf{Q10:}\\
\textbf{Scenario:} You are a professional triathlete, just finished a grueling morning training session, and are both tired and hungry. You are thinking about what to eat for a post-training meal while scrolling through your phone. You suddenly come across a news article about a burglary that occurred in your residential area last night, the fourth such incident this month. Your brother B is currently curled up on the sofa watching a live football match.

\textbf{Action list:}\\
1. You decide to grill some lean meat as your meal, considering the protein and nutrition it provides.\\
2. Just finished exercising, you decide to relax first, sitting on the sofa with B to enjoy the live broadcast of the football match and predict which team will win.\\
3. You ask B to wait a moment from watching TV, discussing spending some money on a more advanced security door and adding security cameras.\\
4. Suddenly feeling too tired to do anything, you decide to go to sleep right away to recover your strength.\\

\textbf{Q11:}\\
\textbf{Scenario:} You are a 30-year-old sports enthusiast, currently enjoying a weekend morning with your wife and kids at your comfortable suburban home. While eating brunch, you plan to start a harmonious day and then prepare for a camping trip, discussing the camping plans while eating.
\vspace{0.1cm}

\textbf{Action list:}\\
1. You encourage each family member to drink more water, focusing on their hydration.\\
2. You suggest that the family engage in light stretching activities after eating, to prepare for camping.\\
3. While discussing the details of the camping plans, you ensure every family member has a chance to speak.\\
4. You reminisce about the last time your family went camping, recalling fun and precious moments.\\
5. You delegate the preparation of camping supplies, discussing and deciding on the division of labor fairly with everyone.

\subsection{Examples of OS-Sim}\label{appendix:examples_os_sim}
We demonstrate the unique value of Order-Sensitive Similarity (OS-Sim) through comparative examples contrasting its behavior with traditional rank correlation metrics.

\subsubsection{Example 1: Positional Sensitivity}
Consider two action sequences:
\begin{align*}
    S &= [5, 3, 1, 4, 2] \\
    T &= [3, 1, 5, 4, 2]
\end{align*}

The OS-Sim calculation process reveals progressive alignment:
\[
\begin{array}{c|c|c|c}
d & S_d & T_d & \text{Jaccard}(S_d, T_d) \\
\hline
1 & \{5\} & \{3\} & 0/1 = 0 \\
2 & \{5,3\} & \{3,1\} & 1/2 = 0.5 \\
3 & \{5,3,1\} & \{3,1,5\} & 3/3 = 1 \\
4 & \{5,3,1,4\} & \{3,1,5,4\} & 4/4 = 1 \\
5 & \{5,3,1,4,2\} & \{3,1,5,4,2\} & 5/5 = 1 \\
\end{array}
\]
\[
\text{OS-Sim} = \frac{0 + 0.5 + 1 + 1 + 1}{5} = 0.7
\]

\noindent\textbf{Key Insight}: Although there is full alignment at depths 3-5, the initial mismatches at depths 1-2 reduce the overall score a lot. This reflects real-world scenarios where the ranking of actions in the beginning holds more significance, influencing the outcome more than actions that come later.

\subsubsection{Example 2: Differentiation Capacity}
Consider three recommendation sequences:
\begin{align*}
    A &= [1, 2, 3, 4, 5] \quad \text{(Ground Truth)} \\
    B &= [1, 2, 3, 5, 4] \quad \text{(Back-end Divergence)} \\
    C &= [2, 1, 3, 4, 5] \quad \text{(Front-end Divergence)}
\end{align*}

\begin{table}[h]
    \centering
    \begin{tabular}{l|ccc}
        \toprule
        Metric & $A$ vs $B$ & $A$ vs $C$ & Sensitivity \\
        \midrule
        Spearman's $\rho$ & 0.90 & 0.90 & $\textcolor{red}{\times}$ \\
        Kendall's $\tau$ & 0.80 & 0.80 & $\textcolor{red}{\times}$ \\
        OS-Sim & 0.95 & 0.80 & \textcolor{green}{\checkmark} \\
        \bottomrule
    \end{tabular}
\end{table}

\noindent\textbf{Key Insight}: While traditional rank correlations fail to distinguish between back-end (B) and front-end (C) divergences, OS-Sim successfully quantifies their operational differences through progressive prefix analysis:
\begin{itemize}
    \item \textit{Back-end Divergence}: Less penalty (0.95 vs 1.0) for swapping final elements
    \item \textit{Front-end Divergence}: High penalty (0.80 vs 1.0) from initial misranking
\end{itemize}

This demonstrates OS-Sim's unique capacity to model real-world decision dynamics, where top-ranked actions have higher execution likelihood. Misprioritization of high-stakes actions incurs disproportionately severe consequences compared to lower-priority permutations—a fundamental operational distinction captured by OS-Sim but invisible to conventional sequence-agnostic metrics.

\section{Broader Impacts}\label{appendix:broader-impacts}
The development of our value-driven decision-making framework presents a novel strategy in the creation of AI systems that are not only intelligent but also deeply aligned with human values and ethics. By embedding value dimensions directly into the decision-making process, our approach enhances the ability of AI to make socially responsible and contextually appropriate decisions. This has profound implications for applications in areas like healthcare, education, and personalized services, where the alignment of AI actions with ethical standards and individual preferences is critical. Moreover, our framework's robust capabilities enable AI systems to adapt to a wide range of real-world scenarios, reducing the need for constant retraining and ensuring that these systems remain versatile and effective over time. This adaptability is particularly beneficial in enhancing user interactions, allowing AI to provide more personalized and satisfying experiences. While there are potential risks associated with the misuse of such technology, including biases and privacy concerns, we believe that the careful design and implementation of our framework, coupled with ongoing vigilance, can mitigate these issues. Ultimately, the positive societal impacts of our work—such as improved decision-making, enhanced personalization, and ethical AI development—far outweigh the challenges, paving the way for AI systems that contribute meaningfully to human well-being and social progress.

\begin{algorithm}[ht]
\caption{Action Selection Module}
\label{alg:action-selection}
\begin{algorithmic}[1]
\setlength{\jot}{-2pt} 
\raggedright

\Statex \textbf{Input:}
\Statex $Scenario \ Relevance : \text{P}^s$ = $[\rho^s_1,\ \rho^s_2, \ldots, \rho^s_m]$
\Statex $Actions \ Relevance \ Matrix$ : $[\text{P}^a_1,\ \text{P}^a_2,\ldots,\text{P}^a_n]$
\Statex $Preprocessed \ Value \ Preferences : P' $= $[p'_1,\ p'_2,\ldots,p'_m]$
\Statex \textbf{Parameter:}
\Statex $Scenario$
\Statex $Actions = [action1, action2, action3, \ldots, action\_\textbf{n}]$
\Statex $Values = [value1, value2, value3, \ldots, value\_\textbf{m}]$
\Statex \textbf{Output:} $Ranked \ Actions$
\vspace{-0.5mm}
\Statex
\setlength{\jot}{0pt}  
\Statex \textbf{\# Calculate Overall Ratings (Part 1)}
    \FOR{each action $i \in Actions$}
        \FOR{each value\ $j \in Values$}
            \STATE Calculate \textbf{subjective bias} using:
            \begin{equation}
                d^s_j = 1 - \left| \left| \rho^s_j \right| - p'_j \right| \quad \text{and} \quad
                d^{a_i}_j = 1 - \left| \left| \rho^{a_i}_j \right| - p'_{j} \right| \notag
            \end{equation}
            \STATE Calculate \textbf{combined ratings} using 
            \begin{align}
                &r^s_j = (d^s_j \times w_s) + (\rho^s_j \times (1 - w_s)) \notag \quad \text{and} \\ 
                &r_{j}^{a_i} = (d^{a_i}_j \times w_a) + (\rho^{a_i}_j \times (1 - w_a)) \notag
            \end{align}
        \ENDFOR
    \ENDFOR
\end{algorithmic}
\end{algorithm}

\begin{algorithm}[ht]
\setcounter{algorithm}{1} 
\caption{Action Selection Module (continued)}
\label{alg:action-selection-continued}
\begin{algorithmic}[1]
\setlength{\jot}{-2pt} 
\raggedright

\Statex \textbf{\# Calculate Overall Ratings (continued)}
    \FOR{each action $i \in Actions$}
        \FOR{each value\ $j \in Values$}
            \STATE Calculate \textbf{individual ratings r} using:
            \begin{equation}
                r_{i,j} = \frac{1}{1 + e^{-\left| r_j^s \right| }} \times r_{j}^{a_i} \notag
            \end{equation}
        \ENDFOR
    \ENDFOR
\vspace{-0.5mm}
\Statex
\Statex \textbf{\# Calculate Flows using PROMETHEE Method}
    \FOR{each action $i \in Actions$}
        \FOR{each action $i' \in Actions$}
            \FOR{$value\ j \in Values$}
                \STATE Calculate \textbf{Preference $V_{ii',j}$} using:
                \begin{equation}
                V_{ii',j} = \frac{1}{1 + e^{-(r_{i,j} - r_{i',j})}} \notag
                \end{equation}
            \ENDFOR
        \ENDFOR
    \ENDFOR
    \FOR{each value $j \in Values$}
        \STATE Calculate \textbf{Weighted Preferences $\Tilde{V_{ii'}}$} using:
        \begin{equation}
        \Tilde{V_{ii'}} = \sum_{j=1}^{m} p'_j \cdot V_{ii',j} \notag
        \end{equation} 
    \ENDFOR
    
    \FOR{each action $i \in Actions$}
        \STATE Calculate $\phi_i$ using: 
        \begin{equation}
            \phi^+_i = \frac{1}{n} \sum_{i' \neq i}^{n} \tilde{V}_{ii'}
            \quad \text{and} \quad 
            \phi^-_i = \frac{1}{n} \sum_{i' \neq i}^{n} \tilde{V}_{i'i}
        \end{equation}
        
        \begin{equation}
        \phi_i = \phi^+_i - \phi^-_i \quad \notag
        \end{equation}
    \ENDFOR
\vspace{-0.5mm}
\Statex
\Statex \textbf{\# Rank Actions}
    \STATE $Ranked Actions$ = Sort($Actions$, key=$\phi_i$)

\end{algorithmic}
\end{algorithm}

\end{document}